\documentclass[10pt,twocolumn,letterpaper]{article}

\usepackage[pagenumbers]{iccv} %

\usepackage{float}
\usepackage{graphicx}
\usepackage{bbding}
\usepackage{pgfplots}
\usepackage{algorithm}
\usepackage{algpseudocode}
\usepackage{dsfont}
\usepackage{comment}
\usepackage{fix-cm}
\usepackage{ragged2e}
\newcommand{\tablesize}{\fontsize{7}{10}\selectfont}
\pgfplotsset{width=10cm,compat=1.9}
\algrenewcommand\algorithmiccomment[1]{\hfill\textcolor{blue}{$\triangleright$ #1}}

\DeclareMathOperator*{\argmin}{arg\,min}
\DeclareMathOperator*{\argmax}{arg\,max}
\DeclareMathOperator*{\topk}{TopK}
\DeclareMathOperator{\concat}{Concat}
\DeclareMathOperator{\onehot}{OneHot}
\DeclareMathOperator{\addNoise}{AddNoise}
\DeclareMathOperator{\SO}{SO}
\DeclareMathOperator{\SE}{SE}

\newcommand{\duster}{DUSt3R }

\newcommand{\lp}{\left(}
\newcommand{\rp}{\right)}
\newcommand{\xB}{\mathbf{x}}

\newcommand{\tB}{\mathbf{t}}
\newcommand{\FB}{\mathbf{F}}

\newcommand{\KB}{\mathbf{K}}

\newcommand{\XB}{\mathbf{X}}

\newcommand{\aB}{\mathbf{a}}

\newcommand{\RB}{\mathbf{R}}

\newcommand{\CB}{\mathbf{C}}
\newcommand{\VB}{\mathbf{V}}
\newcommand{\WB}{\mathbf{W}}
\newcommand{\pB}{\mathbf{p}}
\newcommand{\real}{\mathbb{R}}
\newcommand{\integer}{\mathbb{N}}
\newcommand{\eye}{\mathbb{I}}
\newcommand{\binary}{\{0, 1\}}
\newcommand{\mesh}{\mathcal{M}}

\newcommand{\rankonecolor}{\cellcolor{ForestGreen!70}}
\newcommand{\ranktwocolor}{\cellcolor{ForestGreen!37}}
\newcommand{\rankthreecolor}{\cellcolor{ForestGreen!10}}

\newcommand{\uvec}[1]{\boldsymbol{\hat{\textbf{#1}}}}

\definecolor{iccvblue}{rgb}{0.21,0.49,0.74}
\usepackage[pagebackref,breaklinks,colorlinks,allcolors=iccvblue]{hyperref}

\newcommand{\affspace}{\hspace{2em}}

\title{Geometry-Aware Diffusion Models for Multiview Scene Inpainting}

\author{
Ahmad Salimi\textsuperscript{1}
\and
Tristan Aumentado-Armstrong\textsuperscript{1,3}
\and
Marcus A.~Brubaker\textsuperscript{1,2,4} 
\and
\vspace{+10pt} Konstantinos G.~Derpanis\textsuperscript{1,2,3} \\
{\normalsize \textsuperscript{1}York University \affspace
\textsuperscript{2}Vector Institute for AI \affspace
\normalsize\textsuperscript{3}Samsung AI Centre Toronto \affspace
\textsuperscript{4}Google DeepMind}\\
{\tt\small \{ahmadsa,marcus.brubaker,kosta\}@yorku.ca, tristan.a@samsung.com}\\
}

\begin{document}
\maketitle
\begin{abstract}

In this paper, we focus on 3D scene inpainting, where parts of an input image set, captured from different viewpoints, are masked out. The main challenge lies in generating plausible image completions that are geometrically consistent across views. Most recent work addresses this challenge by combining generative models with a 3D radiance field to fuse information across a relatively dense set of viewpoints. However, a major drawback of these methods is that they often produce blurry images due to the fusion of inconsistent cross-view images.  To avoid blurry inpaintings, we eschew the use of an explicit or implicit radiance field altogether and instead fuse cross-view information in a learned space.  In particular, we introduce a geometry-aware conditional generative model, capable of multi-view consistent inpainting using reference-based geometric and appearance cues. A key advantage of our approach over existing methods is its unique ability to inpaint masked scenes with a limited number of views (i.e., few-view inpainting), whereas previous methods require relatively large image sets for their 3D model fitting step.
Empirically, we evaluate and compare our scene-centric inpainting method
on two datasets, SPIn-NeRF and NeRFiller, which contain images captured at narrow and wide baselines, respectively, and achieve state-of-the-art 3D inpainting performance on both. Additionally, we demonstrate the efficacy of our approach in the few-view setting compared to prior methods.
Our project page is available at \href{https://geomvi.github.io/}{https://geomvi.github.io}.\end{abstract}

\vspace{-2em}\section{Introduction}
\label{sec:intro}

Image inpainting is a long-standing problem in computer vision and graphics \cite{bertalmioimage}.
Unlike unconditional generation, inpainting is constrained by the conditioning input, requiring a visually-plausible output that matches the partial content.
The problem of inpainting \textit{3D scenes} has recently grown in popularity (e.g., \cite{spinnerf,nerf.in}) due to the advent of powerful novel view synthesis (NVS) models.
Such models are usually implemented as scene models capable of differentiable rendering, e.g., neural radiance fields (NeRFs) \cite{original.nerf} or 3D Gaussian splatting (3DGS) \cite{kerbl3Dgaussians}.
For both NVS and 3D inpainting, the input scene is implicitly represented through a posed set of images; hence, we may consider the equivalent problem of ``multiview inpainting'', which provides a natural interface between 2D image inpainting and 3D NVS.
     
\begin{figure}[t]
    \centering
    \includegraphics[width=\linewidth]{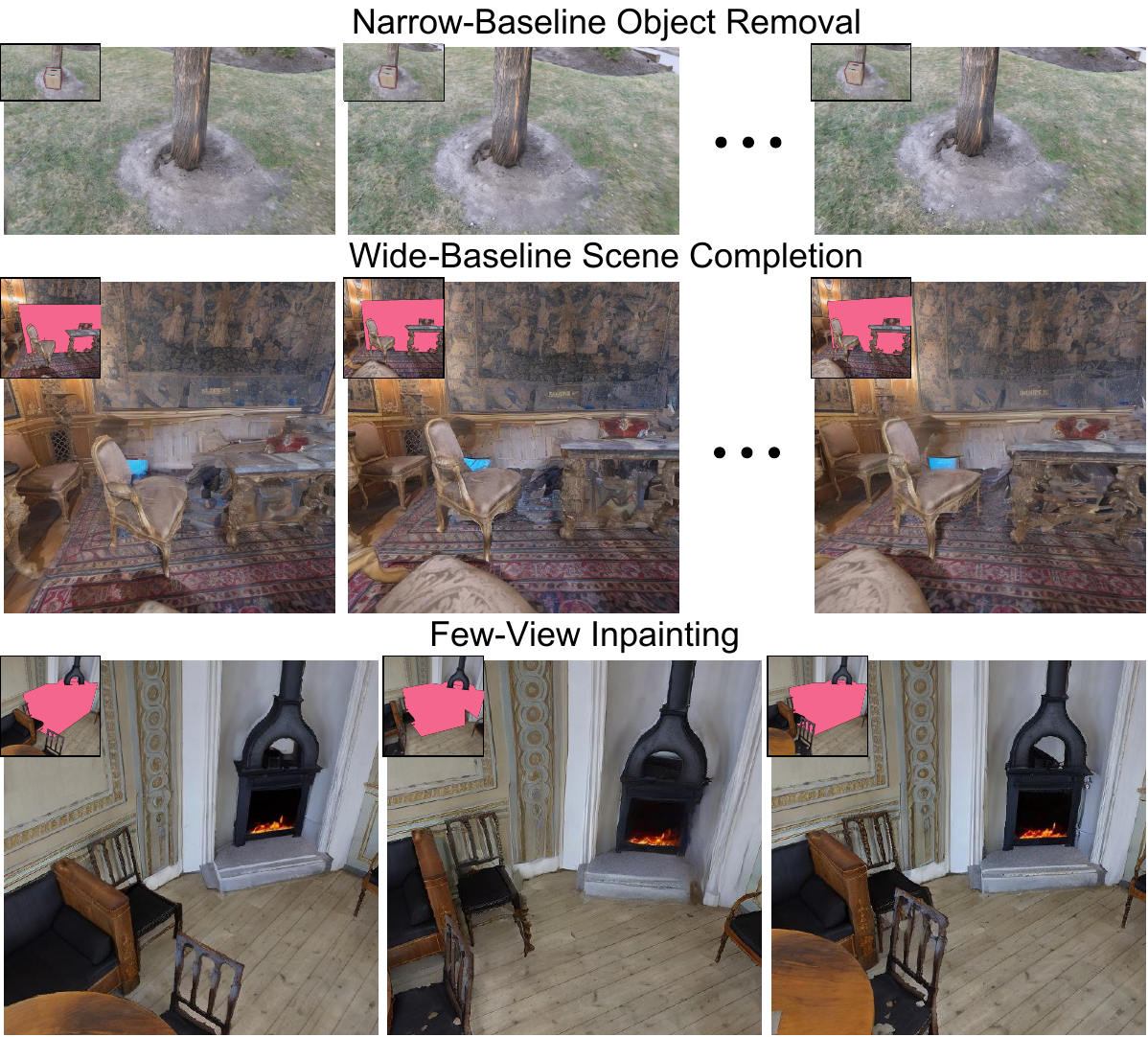}
    \vspace{-15pt}
    \caption{Visualization of our target inpainting tasks.
    We target three tasks: (i) narrow-baseline object removal,
    (ii) wide-baseline scene completion and (iii) few-view inpainting.
    Here, we show examples of our outputs for each task, with
    the corresponding masked inputs shown in the top left corner of each image.
    }
    \label{fig:teaser}
    \vspace{-15pt}
\end{figure}

Multiview (3D) inpainting is significantly more constrained than even the 2D case: inpainted scene content must be \textit{geometrically} realistic and exhibit \textit{cross-view consistency}.
Cross-view consistency is particularly critical when leveraging 2D image inpainters that, by default, are neither aware of the 3D scene structure nor the current state of inpainted content in other images -- unsurprisingly, this results in inconsistent scene content. 
Yet, we still seek to exploit the powerful generative priors learned by 2D inpainters, as the paucity of 3D scene data prevents training inpainters that operate directly in 3D to the same level of quality.
This problem is likely exacerbated with increasingly powerful generative inpainters which tend to hallucinate even more aggressively \cite{reference.guided.nerf}.

A common approach, often used in NeRF editing (e.g., \cite{weber2024nerfiller}), is to fuse information across views, via 3D
radiance fields.
A cyclic process, such as iterative dataset update (IDU) \cite{in2n}, is then used, whereby images are edited, used to fit the NVS model, and NVS renders (which combine information across views) assist with new edits.
While this approach ensures consistency via the
3D radiance field, it has a few shortcomings: 
(i) a tendency towards blurriness, partly due to the fusion happening in pixel space, and
(ii) reliance on the radiance field, which requires accurate camera parameters with sufficient view coverage.

In this work, we resolve these problems by 
fusing cross-view information in a learned space, during a generative diffusion process, 
and eschewing the necessity of a 3D radiance field, though we can optionally choose to use one as a separate post-fitting step.
We further reduce the tendency of generative inpainters to over-hallucinate, a common cause of 3D inconsistencies, by conditioning them on the 3D scene structure.
In particular, we devise a geometry-aware conditional diffusion model, capable of inpainting multiview-consistent images based on geometric and appearance cues from reference images.
One key capability of our model is the handling of \textit{uncertain} or \textit{partial} scene information (e.g., missing due to occlusions and viewpoint changes).
We integrate our model into a 3D scene inpainting algorithm, performing 3D-consistent propagation of image content across views. 
Unlike NeRF-based inpainters, which fuse inconsistencies in appearance space and thus induce blur, our method fuses cross-view information via the generative model, resulting in sharper outputs even when a NeRF is fit to our final inpainted results. 
In this paper, we target three main inpainting tasks: narrow-baseline object removal,
wide-baseline scene completion, and few-view inpainting.
Most previous work has been focused on the first task, the second one
is a recent addition, and the third task has not been 
extensively explored with recent innovations. \cref{fig:teaser} provides a visualization of each of our target tasks.
We evaluate our multiview inpainter on two datasets, SPIn-NeRF \cite{spinnerf} and NeRFiller \cite{weber2024nerfiller}, which contain narrow and wide baselines, respectively, and achieve state-of-the-art 3D inpainting performance on both.
Please see our project page for visualizations of results: \href{https://geomvi.github.io/}{https://geomvi.github.io}.

\section{Related Work}
\label{sec:related-work}

\noindent\textbf{2D Priors for 3D Inpainting.}
2D inpainting has been widely explored \cite{quan2024inpaintingsurvey}. Current state-of-the-art methods leverage conditional diffusion models \cite{stable.diffusion,zhang2023controlnet,ju2024brushnet}, which exhibit high-quality and diverse generations. However,
direct application of independent 2D inpainters for 3D inpainting
leads to 3D inconsistencies \cite{spinnerf,reference.guided.nerf}. 
Yet, despite the challenges in multiview inconsistency, recent methods \cite{weber2024nerfiller,prabhu2023inpaint3d,mirzaei2024reffusion,chen2024mvip,lin2025taming} have explored combining diffusion-based 2D inpainters with explicit or implicit 3D radiance fields to exploit their powerful generative prior.
To preserve this prior while enforcing greater 3D consistency, we further constrain a diffusion-based inpainter using scene geometry.

\noindent\textbf{3D Scene Editing.} Editing 3D %
scenes is essential for 3D content creation (e.g., for video games or virtual reality).
Spurred by the rise of
3D radiance fields
(e.g., \cite{rabby2023beyondpixels,chen2024survey}) there has been an explosion of techniques.
There are numerous forms of 3D editing, including 
scene translation 
(e.g., \cite{in2n,wys,vicanerf,koo2024posterior,chen2024dge,chen2024gaussianeditor,wu2024gaussctrl,wang2025view}),
super-resolution 
(e.g., \cite{huang2023refsr,lee2024disr}),
shape deformation 
(e.g., \cite{yuan2022nerf,zheng2023editablenerf,jambon2023nerfshop}),
appearance alterations
(e.g., \cite{mazzucchelli2024irene,kuang2023palettenerf,wang2023seal,lee2023ice}),
and
inpainting
(e.g., \cite{spinnerf,weder2023removing,nerf.in,reference.guided.nerf}),
which is the focus here.

Many 3D inpainters focus primarily on \textit{object removal} \cite{wang2025learning,lu2024view,spinnerf,weder2023removing,wang2024innerf360}.
In contrast, our method can insert additional content and perform scene completion.
RenderDiffusion \cite{anciukevivcius2023renderdiffusion} enables 3D-aware inpainting, but relies on weak supervision (utilizing only 2D supervision) and is limited to simpler scenes.
SIGNeRF \cite{dihlmann2024signerf} specializes in localized translation and object insertion, rather than general 3D inpainting. 
Chen et al.~\cite{chen2025single} examines the impact of the inpainting mask for deterministic object removal, but is restricted to forward-facing scenes.
Gaussian Grouping \cite{ye2023gaussian} integrates segmentation features directly into the radiance field, facilitating various editing tasks.
For general inpainting, most methods build on 3D radiance fields
\cite{chen2024mvip,mirzaei2024reffusion,prabhu2023inpaint3d,liu2024infusion,lin2025taming}, with several approaches (e.g., \cite{mirzaei2024reffusion,prabhu2023inpaint3d,chen2024mvip})
leveraging score distillation sampling (SDS) \cite{poole2022dreamfusion,wang2023score} to incorporate 2D diffusion priors.
However, these methods typically require a relatively \emph{dense view coverage} for 3D model fitting.
In contrast, our method inpaints the image set directly, allowing it to operate effectively even with sparse view coverage.
Similar to our work, several recent inpainters also alter diffusion models, 
specializing them for multiview inpainting \cite{cao2024mvinpainter,mirzaei2024reffusion}.
For our evaluation, we utilize the scene-centric settings of
NeRFiller \cite{weber2024nerfiller} and SPIn-NeRF \cite{spinnerf},
which covers %
a variety of scene types and camera baselines.

\noindent\textbf{Reference-Conditioned Generative Editing.}
A variety of methods have been devised to encourage consistency across multiple generations
(e.g., \cite{tewel2024training,zhou2024storydiffusion,avrahami2024chosen,sajnani2024geodiffuser}), 
usually by sharing features across diffusion processes.
However, these methods only enforce \emph{semantic} consistency, which is insufficient for precise, pixel-level coherence required for 3D-consistent content.
Separately, conditional image generators can perform NVS
by mapping observed reference images and target camera parameters to a new view (e.g., \cite{liu2023zero,shi2023zero123++,yang2024consistnet,gao2024cat3d,yu2024polyoculus,tseng2023consistent,yu2023long,chan2023generative,yu2024viewcrafter}) by training on multiview data.
However, these methods are meant for generative NVS, rather than inpainting existing 3D scenes coherently.
While our conditional diffusion model is reminiscent of this, addressing the entirety of the NVS problem is not necessary for 3D inpainting; instead, we assemble reference information (from multiple source views) in the image coordinate frame of the target view (i.e., no camera information is sent to the model).
Hence, our inpainter avoids learning complex 3D transforms (e.g., triangulating and projecting between frames). Instead, it learns to utilize projected reference information entirely in 2D, while accounting for simple 3D constraints, like relative depth ordering of generated content with respect to existing content.
Finally, ``reference-based inpainting'' \cite{zhao2023geofill,zhao20223dfill,zhou2021transfill} uses one reference image to inpaint another, sometimes handling transforms beyond viewpoint, like lighting changes. While our method also leverages reference information, it aligns more with NeRF-based approaches, operating on full multiview image sets where all views are initially masked, rather than a single reference-target pair.

\noindent\textbf{Iterative 3D Generation.} 
3D content generation is typically iterative or autoregressive. Iterative methods,
common in text-to-3D \cite{poole2022dreamfusion,mcallister2024rethinkingsds} and instruction-based 3D editing \cite{in2n}, have also been used for 3D inpainting via SDS \cite{poole2022dreamfusion} (e.g., \cite{prabhu2023inpaint3d,mirzaei2024reffusion,chen2024mvip}) and IDU \cite{in2n} (e.g., \cite{weber2024nerfiller}).
In contrast, autoregressive approaches have been explored in 3D scene generation (e.g., \cite{chung2023luciddreamer,hoellein2023text2room}) and generative NVS (e.g., \cite{yu2023long,liu2021infinitenature,rombach2021geogpt,ren2022lookout,tseng2023consistent}). We introduce an autoregressive approach for inpainting large-baseline scenes.

\begin{figure*}[!ht]
    \centering
    \includegraphics[width=\linewidth]{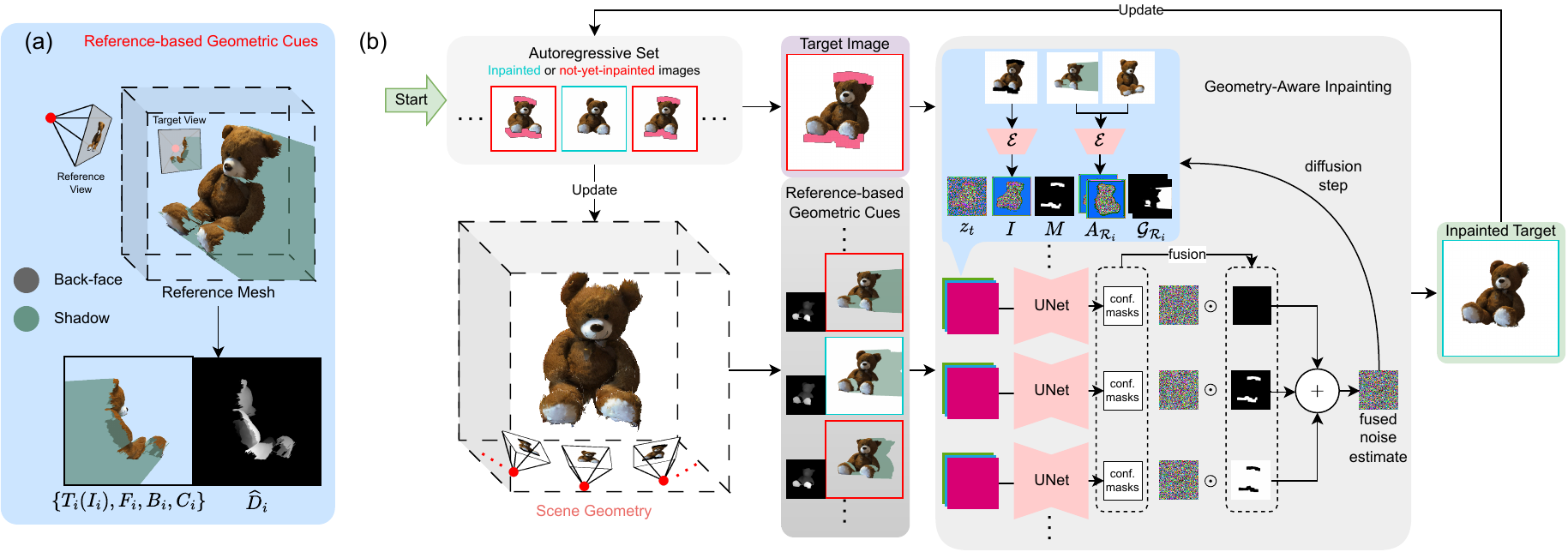}
    \vspace{-20pt}
    \caption{
    Overview of our geometric-aware 3D scene inpainter.
    (a) A visualization of the reference-based geometric cues. Back-faces are always covered by shadow volumes. $T_i(I_i), F_i, B_i, C_i, \widehat{D}_i$ denote the rendered photometric content, front-face mask, back-face mask, shadow mask, and disparity, respectively, for the reference view $i$.
    (b) A step-by-step visualization of our autoregressive inpainting process. Note that the scene geometry consists of separate meshes for each image, not a single harmonized mesh, as shown here for 
    simplicity.
    Here, we are only showing one diffusion step for the geometry-aware inpainting model. $\mathcal{E}, z_t, I, M, A_{\mathcal{R}_i}, \mathcal{G}_{\mathcal{R}_i}$ denote the VAE encoder (which maps an image to the latent space of Stable Diffusion), the diffusion latent at timestep $t$, the masked image latent, the mask, the appearance cues for reference view $i$, and the geometric cues for reference view $i$, respectively; see \S\ref{supp:subsec:autoregressive-qualitative} for an example of autoregressive steps.
    }
    \vspace{-15pt}
    \label{fig:autoregressive}
\end{figure*}

\section{Method}
\label{sec:method}

\noindent\textbf{Setting.}
As with prior work \cite{spinnerf}, we assume as given 
a set of $N$ views, $I_i \in \real^{H \times W \times 3}$, representing a static scene, and 
their corresponding inpainting masks, $M_i \in \binary^{H \times W}$, demarcating the 3D region to be inpainted.
While other methods may require additional inputs, such as camera parameters or depths per image, these are optional for our approach (though we can nonetheless use them).
Our objective is to jointly inpaint the given views, thus inpainting the 3D scene, ideally in a consistent manner across views.

\noindent\textbf{Overview.}
Prior works (e.g., \cite{spinnerf,reference.guided.nerf,weber2024nerfiller,prabhu2023inpaint3d,mirzaei2024reffusion,chen2024mvip}) often have two disparate components: 
(i) 
an implicit or explicit 3D radiance field (e.g., NeRF \cite{original.nerf} and 3DGS \cite{kerbl3Dgaussians}),
which fuses information across views to ensure consistency, and
(ii) a 2D image inpainter, which alters the source views, potentially conditioned on the state of (i) (e.g., \cite{in2n,wys,weber2024nerfiller}).
However, this separation has shortcomings:
first, (i) fuses information in pixel space (due to the supervision mechanism), leading to blurry results, and second, (ii) is only aware of the other views through (i), meaning
any mechanism for enforcing consistency is \textit{indirect}.
In contrast, our algorithm relaxes the need
for a radiance field,
fuses information in the \textit{learned} space of a generative model (avoiding the blur of pixel space), and enables \textit{direct}
appearance transfer
across views, using estimated scene geometry.
To attain this, our approach consists of two models: 
a scene geometry estimator and a geometry-aware inpainting model.
For the former, we use the performant \duster \cite{dust3r}, which efficiently provides dense depth, with or without the presence of camera poses, utilizing views (inpainted or not) directly.
For the latter, we fine-tune a latent 2D diffusion-based inpainter, to condition on the other views. However, naively conditioning the inpainter forces it to
learn both inpainting and generative NVS
(itself a non-trivial task \cite{yu2024polyoculus,gao2024cat3d,tewari2023diffusion}),
by learning to map one view to another using an internal scene model.
Instead, we use our scene geometry estimator to feed appearance information from other views to our inpainter, by directly projecting information from source views, as well as passing explicit geometric information pertinent to the inpainting (e.g., occlusion). 

Given these two models, we devise a simple autoregressive scene inpainting algorithm.
At each iteration, we inpaint a subset of not-yet-inpainted views, conditioned on the other views (whether inpainted or not), followed by updating the estimated geometry.
This cyclic process gradually fills in the scene's geometry and appearance;
see \cref{fig:autoregressive} for a schematic of our approach.

For the remainder of this section, we provide details on our approach.
We first briefly review our scene geometry estimator, based on \duster \cite{dust3r} (\S\ref{subsec:sge}).
Next, we describe our geometry-aware inpainting diffusion model in \S\ref{subsec:geo-aware-diffusion}, including the reference-based geometric cues guiding each view's inpainting.
Finally, we present the autoregressive procedure used to iteratively inpaint the scene (\S\ref{subsec:autoregressive}). \S\ref{supp:sec:method}
provides additional methodological details.

\subsection{Scene Geometry Estimation}
\label{subsec:sge}
We utilize \duster \cite{dust3r} for scene geometry reconstruction.
\duster can efficiently estimate scene depths \textit{and} camera parameters from multiview image sets. As we inpaint the scene, we iteratively reapply it to update the scene geometry throughout the process; see \S\ref{supp:subsec:dust3r} for details.

\subsection{Geometry-aware Inpainting Diffusion Model}
\label{subsec:geo-aware-diffusion}

Our goal is to devise an image inpainting model, conditioned on a multiview image set and scene geometry.
The view set may
include both
masked (non-inpainted) and complete (inpainted) images,
used to inpaint the target image.
Thus, this setup is a form of reference-based inpainting, with the additional constraints of a 3D scene structure.

\noindent
\textbf{Notation.}
Formally, let $I$ be the target image to be inpainted
(i.e.,
inpainting one
image), with mask, $M$,
and camera parameters, $\Pi$ (i.e., extrinsics and intrinsics).
Each element of the reference view-set, $\mathcal{R}=\{ (I_i,M_i,b_i,\Pi_i,D_i)\}_i$, consists of an image ($I_i$), mask ($M_i$), indicator of whether $I_i$ has been inpainted
($b_i$), camera parameters ($\Pi_i$), and depth map ($D_i$).
When $b_i = 1$,
we simply set $M_i$ to nullity (i.e., all parts of $I_i$ are trustworthy).
Camera and depth information
come from our geometry estimator (\S\ref{subsec:sge}).
Thus, based on the current state of the scene, our inpainter, $f$, obtains $\widehat{I} = f(I,M|\mathcal{R})$, 
which can 
later
be
used in the reference view-set of future inpaintings (see \S\ref{subsec:autoregressive}).

\subsubsection{Coordinate-aligned Conditional Inpainting.}
\label{subsubsec:cond-inp}
Our conditioning approach leverages geometric knowledge of the scene.
In particular, \textit{%
using scene geometry, we project appearance and geometric information from references into the target camera's image plane};
this alleviates the need for the network to learn geometry estimation and view transforms (as in NVS), saving its capacity for inpainting.
Given depth and cameras, from \S\ref{subsec:sge}, this is a simple
projective mapping.
However, three issues remain: 
(i) selecting cues,
(ii) fusing cues across views, and
(iii) training the inpainter to handle errors in estimated scene geometry.
We first consider (i), then discuss (ii) in \S\ref{subsec:mrcond} and (iii) in \S\ref{subsec:inptrain}.

\noindent\textbf{Preliminaries.}
For each reference image, $I_i$, let $T_i$ be the projective transform from the $i$th camera to the target frame ($I$).
This is performed by constructing a mesh, $S_i$, from $D_i$,
assigning cues as nodal attributes,
and rendering $S_i$ into the target frame via $\Pi$ (see \S\ref{supp:subsec:mesh} for details).

\noindent\textbf{Appearance Cues.}
We provide the inpainter, $f$, with two appearance cues:
(a) direct reference pixels and (b) a stylistic hint.
For (a), we
pass $T_i(I_i)$
instead of $I_i$,
giving
the inpainter direct access to view-aligned pixel
colours for localized convolutional processing.
However, with large camera baselines, projected information from distant viewpoints may be insufficient 
(e.g., 
only the back-face triangles of an object are visible).
To ensure a \textit{stylistic} harmony, we introduce (b), an optional ``hint image'', $H$, which is not projected through the scene geometry but rather provides global appearance characteristics.
When used, we set $H$ as the furthest inpainted image from $I$ in $\mathcal{R}$ (see \S\ref{supp:subsec:hint}).
We denote our appearance cues as
$A_{\mathcal{R}} = \{ T_i(I_i) \}_i \cup \{ H \}$.

\noindent\textbf{Geometric Cues.}
We next obtain a set of geometric cues, controlling
the
reliability
of photometric reference content,
illustrated in \cref{fig:autoregressive} (a).
Specifically, per reference, we compute a
(i) front-face mask,
(ii) back-face mask,
(iii) normalized inverse depth (ordering),
and
(iv) shadow mask,
all in the target coordinate frame.
The mesh $S_i$ has front- and back-faces, indicating the validity of photometric content (i.e., the former has valid appearance from $I_i$, but the latter merely implicates the presence of geometry).
Thus, (i), (ii), and (iii) are 
rendered from
$S_i$,
denoted $F_i$, $B_i$, and $\widehat{D}_i = T_i(D_i)$, respectively.
For (iv), $S_i$ implies a ``shadow volume'' \cite{mccool2000shadow,haller2003real} with respect to $\Pi_i$, 
representing
the 3D space \textit{hidden from the reference camera} 
(see \cref{fig:autoregressive} and \S\ref{supp:subsec:shadow} for details),
which 
is rendered
into the target view, written $C_{i}$.
From the target view, any pixel with a camera ray hitting the shadow volume is explicitly uncertain -- there could be content there or not.
These maps
form
our geometric cue set,
$ \mathcal{G}_\mathcal{R} = \{ F_i, B_i, \widehat{D}_i, C_i \}_i $,
another input to our diffusion model.
Importantly, these cues \textit{also} enable a {hierarchical, confidence-based} fusion of reference information, based on the uncertainty induced by the geometric structure.

\subsubsection{Uncertainty-aware Multireference Conditioning}
\label{subsec:mrcond}

Using our cue-based notation, our method inpaints
an image
$ \widehat{I} = f(I,M|A_\mathcal{R}, \mathcal{G}_\mathcal{R}) $, based on a reference set $\mathcal{R}$.
However, fusing information across references is
challenging,
especially
with
conflicting
content.
We address this by running parallel per-reference diffusion processes and fusing
noise estimates at each diffusion step.
Ideally, this fusion is geometry-aware
(e.g., front-faces of $S_i$ provide
high-certainty
photometric content,
while back-faces only upper bound target view depth).
Since geometric maps
may be slightly misaligned
due to errors in estimated geometry, we alter the diffusion model to predict confidence maps for each reference, 
emulating the aligned geometric masks.
In the following, we describe how our inpainting is implemented, including confidence estimation and 
fusion.

\noindent\textbf{Hierarchical Confidence Estimation.}
The geometric cues, $\mathcal{G}_\mathcal{R}$, 
indicate
the reliable parts of each reference.
We
utilize three geometric signals.
First, the \textit{front-facing confidence mask}, $\CB_f$, indicates a pixel is either outside the inpainting mask 
or guided by a front-facing rendered pixel. In other words, the model has copied the photometric content from the target image itself or the rendered photometric content ($T_i(I_i)$). 
Second, the \textit{back-facing confidence mask}, $\CB_b$, indicates a pixel is geometrically
restricted
by a rendered \textit{back}-face, 
suggesting
new geometry
to
be generated \textit{in front} of it.
Finally, the \textit{shadow confidence mask}, $\CB_s$, signals the model’s certainty in trusting photometric information despite \textit{potential} occlusion (shadow volume).
Notice the confidence masks are closely related to the geometric cues $F_i$, $B_i$, and $C_i$.
Importantly, though, 
these cues
are often slightly misaligned with the actual target image due to geometry estimation errors. 
Thus, we instead modify our diffusion model to \textit{estimate these
confidence masks},
i.e., at every diffusion step, our inpainter not only generates a denoising estimate, but also provides an estimate of these three confidence masks. Please see \S\ref{supp:subsec:training} for details on supervising the confidence masks, and \cref{fig:fusion} for a visualization.

\noindent\textbf{Parallel Diffusion Processing.}
We now formalize the inpainting process.
Given $\mathcal{R}$ and our diffusion model, $f$, 
we split the inpainter into $n=|\mathcal{R}|$ independent streams.
Let 
$ (\varepsilon_{i,t}, \CB_{f,i,t}, \CB_{b,i,t}, \CB_{s,i,t}) = f(I,M|A_i, \mathcal{G}_i) $, 
where $\varepsilon_{i,t}$ is the estimated noise for reference $i$, $A_i = \{H, T_i(I_i)\}$, and $\mathcal{G}_i = \{ F_i, B_i, \widehat{D}_i, C_i \}$, be the output of the $i$th process at
time $t$.
Denote $\mathcal{C}_t = \{\CB_{f,i,t}, \CB_{b,i,t}, \CB_{s,i,t}\}_{i}$
as the combined confidence
maps.
We then fuse the noise estimates,
$ \mathcal{E}_t = \{ \varepsilon_1, \ldots, \varepsilon_n \}$,
to obtain a
fused
estimate, 
$\varepsilon_t = \Gamma(\mathcal{E}_t, \mathcal{C}_t) $,
using our fusion operator, which follows a simple rule-based hierarchy, described in \S\ref{subsubsec:geo-aware-inpainting}.
Before the next timestep, $t-1$, $\varepsilon_t$ is used to update
the noisy latent $z_t$, which is shared
across reference streams.
This fuses multiview information in the \textit{learned, generative} space of the diffusion noise estimate, rather than in pixel space.

\subsubsection{Multiview-aware Training}
\label{subsec:inptrain}

We initialize our
geometry-aware
inpainter with Stable Diffusion v2, fine-tuned for inpainting \cite{stable.diffusion,sdinp}.
Following prior work \cite{ip2p}, we condition on cues by adding zero-initialized channels to the first convolutional layer of the
UNet \cite{u.net}.
We also modify the UNet to output confidence masks alongside noise estimates at each timestep.
For training, 
one
can utilize a multiview dataset (real or synthetic); however, to ensure greater data diversity, we 
also 
\textit{synthesize} data from single-view images, via monocular depth estimation, perturbing the virtual camera, and rendering the starting image as a reference.
To simulate geometry estimation errors, we artificially perturb the reference mesh; in that case, supervision for the confidence masks can be generated via the \textit{un}perturbed mesh; 
see \S\ref{supp:subsec:training} for additional details.

\subsection{Autoregressive Scene Inpainting}
\label{subsec:autoregressive}

With our geometry estimator (\S\ref{subsec:sge}) and geometry-aware inpainter (\S\ref{subsec:geo-aware-diffusion}), we can now iteratively inpaint the entire scene.
To begin, we initialize the scene geometry by computing multi-view metric depth maps of the \textit{in}complete input views. 
We also initialize the ``autoregressive set'' with the incomplete input images, which will be autoregressively inpainted. 
A random view is then selected to start the inpainting. 
Each iteration consists of three steps: 
(i) 
inpainting
a subset of not-yet-inpainted images,
(ii) updating the autoregressive set and the scene geometry with the inpainted images, and 
(iii) selecting a subset of images
to be inpainted
at
the next autoregressive iteration. A high-level illustration of these steps and a step-by-step example are provided in \cref{fig:autoregressive} (b) and \S\ref{supp:subsec:autoregressive-qualitative}, respectively.

\subsubsection{Geometry-aware Inpainting}
\label{subsubsec:geo-aware-inpainting}
For each target view, we select a subset of reference views from the autoregressive set,
prioritizing those already inpainted.
Following \S\ref{subsec:geo-aware-diffusion}, we render appearance and geometric cues from all reference views, run parallel diffusion processes,
and fuse noise estimates at each step via the predicted confidences. 
The fusion, $\Gamma$, follows a four-level confidence hierarchy: (i) front-face confidence, (ii) back-face confidence, (iii) shadow confidence, and (iv) no confidence. For each patch at each level, we select the noise estimate from the closest camera among the views at the same level. This 
ensures
fusion of
the most reliable
reference information
during
denoising;
see \S\ref{supp:subsec:fusion} for additional details.

\subsubsection{Autoregressive Set and Scene Geometry Update}

In this step, we replace the target views in the autoregressive set with their inpainted versions and update their geometry
using the
inpainted images, as detailed in \S\ref{supp:subsec:dust3r}.

\subsubsection{Selecting the Next Images to Inpaint}
\label{subsubsec:next-images}

We employ a two-stage strategy for the autoregressive process. First, we inpaint a wide-baseline subset of the scene, one by one,
to \textit{generate} the missing content of the scene. Then, we \textit{propagate} the generated content to the remaining views simultaneously.
At the start of
the first stage, we use a greedy min-max approach to select the wide-baseline subset. Beginning with the first view, we iteratively add the view that maximizes the minimum distance to the
existing subset.
Once selected, we order them to minimize distance to prior views
(see \S\ref{supp:subsec:wide-baseline} for details).
The wide-baseline stage
follows the
sorted order,
with each target view conditioned on the entire autoregressive set,
inpainted or not.
In the propagation step, 
remaining views
are conditioned only on the inpainted wide-baseline images.

\section{Empirical Evaluation}
\label{sec:experiments}

We evaluate our 3D scene inpainting method and compare it to previous methods across three settings: (i) 
object removal on narrow-baseline scenes,
(ii) scene completion with wide baselines, and (iii) inpainting with few-view inputs.

\subsection{Implementation Details}
We use PyTorch3D \cite{pytorch3d} for mesh rendering. For training,
we use a mixture of two datasets, MS COCO \cite{coco.dataset} and Google Scanned Objects \cite{gso.dataset}. 
For MS COCO, we synthesize the reference-based geometric cues using DepthAnything V2 \cite{depth.anything.v2}.  
Training and evaluation is performed on NVIDIA L40 GPUs (16 for training, one for inference).
Additional implementation details are provided in \S\ref{supp:sec:impl}.

\subsection{Evaluation Protocol}
\label{subsec:eval-protocol}

\noindent\textbf{Datasets.}
To evaluate our method for narrow-baseline inpainting, we use the SPIn-NeRF \cite{spinnerf} dataset, a widely used benchmark for object removal in front-facing real-world scenes. This dataset
contains
10 scenes, each with 60 images 
featuring
the object to be removed
and corresponding masks,
along with
40 images without the object as ground truth for evaluation.
In addition, we also use the scene-centric portion of the NeRFiller \cite{weber2024nerfiller} dataset to further investigate the performance of our method on more complex scenes,
particularly those with larger baselines.
This dataset serves as a benchmark for the scene completion task. 
For the few-view inpainting task, we use SPIn-NeRF and the scene-centric portion of NeRFiller, resulting in a total of 15 scenes. For each scene, we uniformly sample eight subsets of two views and eight subsets of three views. This yields a total of 240 few-view sets. Please see \S\ref{supp:sec:dataset} for details.

\noindent\textbf{Baselines.}
For object removal, we compare our method to state-of-the-art approaches on the SPIn-NeRF dataset, specifically:
SPIn-NeRF \cite{spinnerf}, which inpaints images and depth maps to supervise NeRF fitting;
Inpaint3D \cite{prabhu2023inpaint3d}, which uses SDS \cite{poole2022dreamfusion} to inpaint a NeRF with a diffusion prior;
RefFusion \cite{mirzaei2024reffusion}, which adapts an inpainting diffusion model to a reference image and uses SDS to inpaint a 3DGS \cite{kerbl3Dgaussians} with the reference-adapted diffusion model;
InFusion \cite{liu2024infusion}, which inpaints a 3DGS by inpainting the depth map of an inpainted reference image;
MVIP-NeRF \cite{chen2024mvip}, which uses SDS to inpaint a NeRF by inpainting the rendered images and normal maps via a diffusion prior,
and MALD-NeRF \cite{lin2025taming}, which inpaints a NeRF by performing masked adversarial training for per-scene customization of a diffusion model.
InFusion \cite{liu2024infusion} also evaluates Gaussian Grouping \cite{ye2023gaussian} on SPIn-NeRF, a 3DGS-based segmentation method that does not directly evaluate on SPIn-NeRF itself.
For scene completion, we compare our method to 
Stable Diffusion \cite{stable.diffusion}, which is the naive baseline of independent 2D inpainting, and NeRFiller \cite{weber2024nerfiller}, which 
alternates between editing input images, and updating the 3D representation encompassed by a NeRF
to fuse the edited images, this process is collectively called iterative dataset update. 
Finally, for the few-view task, we compare our method to SPIn-NeRF \cite{spinnerf} and NeRFiller \cite{weber2024nerfiller}. For this task, we use NeRFiller directly.
However, since SPIn-NeRF relies on supervision via COLMAP's \cite{schoenberger2016sfm} sparse depth, we disable this supervision for the few-view task, as scenes from the NeRFiller dataset lack COLMAP information.

\begin{figure*}[t]
    \centering
    \includegraphics[width=\linewidth]{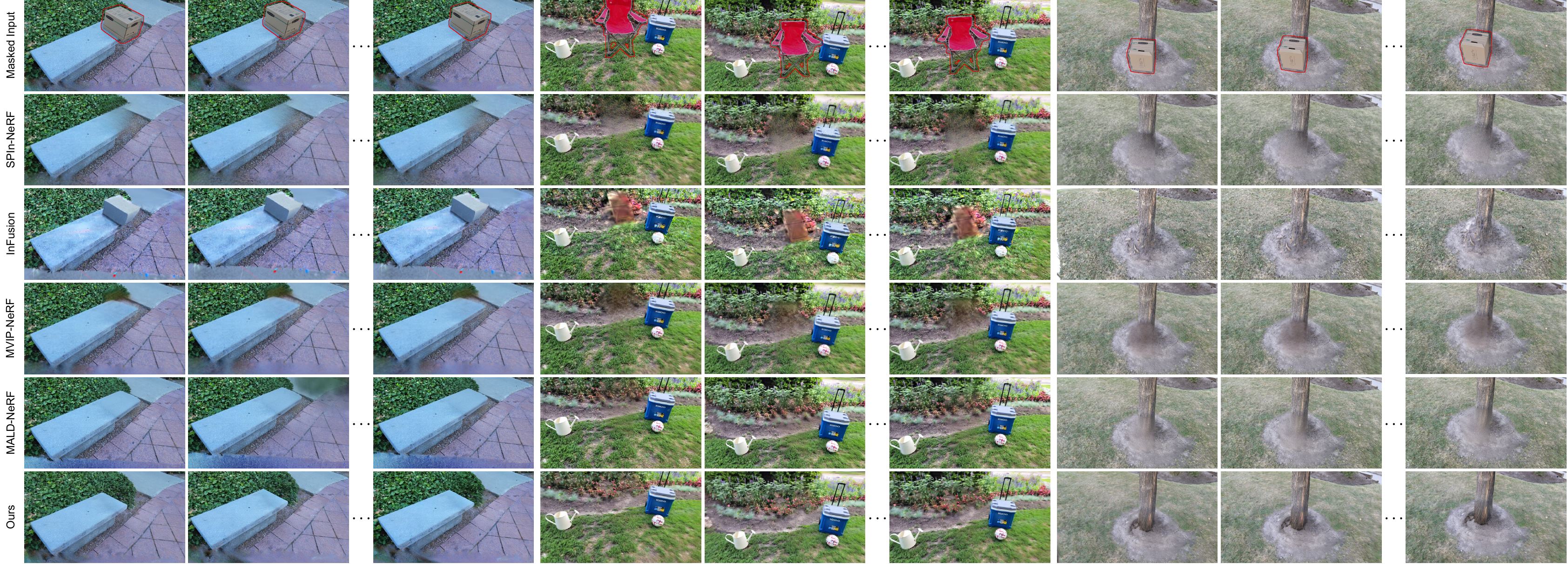}
    \vspace{-20pt}
    \caption{
    Qualitative object removal comparisons on the SPIn-NeRF dataset. Notice other methods produce blurry regions (e.g., bench end) due to multiview inconsistencies, while ours preserves sharpness and visual plausibility. Please zoom in for details.
    }
    \label{fig:spinnerf-qualitative}
    \vspace{-15pt}
\end{figure*}

\noindent\textbf{Metrics.} 
For the SPIn-NeRF dataset, we use the same evaluation 
protocol as reported in the original paper \cite{spinnerf}.
There is no publicly available evaluation code provided with the SPIn-NeRF dataset; we confirmed the details of the protocol with the authors.
Specifically, we compute LPIPS \cite{lpips} (with VGG-16 \cite{vgg}) and FID \cite{fid} between the inpainted images and the ground-truth images, cropped by the inpainting mask's bounding box. The bounding box's size is increased by 10\% before cropping, uniformly in each direction. We additionally assess the sharpness of the inpainted images within the inpainting mask using the Laplacian variance \cite{pertuz2013analysis}.
As our inpainting method does not explicitly enforce 3D consistency
via a 3D radiance field,
we evaluate our consistency using 
the TSED metric \cite{yu2023long}.
Here, our feature correspondences are limited to those inside the inpainting 
masks.
On the SPIn-NeRF dataset, as the scenes have very small baselines, correspondences are considered across all possible view pairings.

For the NeRFiller dataset, we use the same evaluation metrics as NeRFiller \cite{weber2024nerfiller}. As these metrics are computed on NeRF renders, we fit a NeRF on our inpainted images directly. For the image metrics, PSNR, SSIM, and LPIPS, we compare the rendered training views of the fitted NeRF to the inpainted images.
For the video-based metrics, we compute MUSIQ \cite{ke2021musiq} and Corrs on a video rendered from the NeRF. 
We also evaluate the sharpness of both inpainted images and the NeRF renders.
Finally, we compute TSED to evaluate consistency across images.
As scenes in the NeRFiller dataset have a wide baseline, we only consider the two closest views in the view pairs.

Finally, for the few-view inpainting task, as there is no ground truth available, we only compute sharpness and MUSIQ to evaluate image quality. We also compute Corrs and TSED on all possible image pairs in the few-view set. All metrics are computed only inside the bounding box around the inpainting mask for the few-view inpainting task. \S\ref{supp:sec:metrics} provides additional details on computing TSED.

\subsection{Results}

\noindent\textbf{Object Removal on Narrow-baseline Scenes.}
Due to the narrow baseline, we perform single-reference inpainting for this task. As our method does not rely on fitting a NeRF on the training views, and the evaluations are performed on the test views, we follow SPIn-NeRF's \cite{spinnerf} procedure to evaluate image inpainters, e.g., LaMa \cite{lama}. Specifically, we first fit a NeRF on the training views and render the test views, which will now contain the unwanted object. The rendered test views are then used as inputs to our inpainting method.
To evaluate SPIn-NeRF, InFusion\footnote{The reported results for InFusion (arXiv-only preprint) are not qualitatively or quantitatively reproducible (others report similar issues on GitHub).}, and MVIP-NeRF, we run their official code to reproduce the results.
Since RefFusion and Inpaint3D do not provide publicly available code, we report the numbers provided by the papers.
For MALD-NeRF, we evaluate their publicly available inpainted images for the SPIn-NeRF dataset.
As shown in
\cref{tab:spinnerf},
our method outperforms all baselines on the SPIn-NeRF benchmark.
We obtain comparable sharpness and LPIPS to MALD-NeRF, while significantly outperforming it on other metrics
(sharpness and TSED). Further, 
we demonstrate the ability to handle sparse-view inpainting (see below), which cannot be easily handled by NeRF-based approaches.
Please see \S\ref{supp:subsec:maldnerf-artifacts} for further analysis.
Our method is also efficient, achieving faster scene inpainting compared to other methods.
Furthermore, 
the TSED results
show that despite other methods utilizing an explicit or implicit radiance field to enforce 3D consistency, our method achieves a higher consistency score, primarily due to blurry outputs. Please see \S\ref{supp:sec:tsed-eval} for a comprehensive TSED analysis. 
Finally, we present a set of qualitative results in \cref{fig:spinnerf-qualitative}, 
showing
our method produces sharper images 
than baselines,
indicating the efficacy of our cross-view fusion, which operates in a \textit{learned} space, rather than pixel space.

\begin{table}[t]
\centering
\tablesize
\begin{tabular}{c|ccc|c|c}
Method                               & LPIPS $\downarrow$             & FID $\downarrow$               & $\sigma \uparrow$ & $T_{2\text{px}} \uparrow$ & $\tau \downarrow$       \\ \hline
Inpaint3D
\cite{prabhu2023inpaint3d}  & 0.5150                         & 226.04                         & -                                       & -                           \\
\hline
InFusion \cite{liu2024infusion}                           & 0.4210                         & 92.62                          & -                                       & -                           & -                         \\
Gaussian Grp. \cite{ye2023gaussian} & 0.4540 & 123.48 & - & - & - \\
\hline
SPIn-NeRF
\cite{spinnerf}          & 0.4864                         & 160.42                         & \rankthreecolor13.74                    & \ranktwocolor61.04  
 & \rankthreecolor1h 40m                      \\
RefFusion
\cite{mirzaei2024reffusion} & \rankthreecolor0.4283          & -                              & -                                       & -     & -                      \\
InFusion
\cite{liu2024infusion}
& 0.6692                & 244.19                & 9.30                 & 35.88                  & \ranktwocolor14m      \\
MVIP-NeRF
\cite{chen2024mvip}      & 0.5268                         & 215.60                              & 11.96                                       & \rankthreecolor58.33 & 17h 38m                           \\
MALD-NeRF
\cite{lin2025taming}    & \rankonecolor0.3996  & \ranktwocolor130.95 & \rankonecolor35.27 & 58.22 & -   \\
Ours                                                      & \ranktwocolor0.4028 & \rankonecolor108.36 & \ranktwocolor34.50  & \rankonecolor67.35 & \rankonecolor12m
\end{tabular}\\
\justifying
\vspace{-5pt}
\caption{Quantitative evaluation of the object removal task on SPIn-NeRF dataset.
We denote sharpness ($\times 10^{-5}$) as $\sigma$, the percentage of consistent image pairs (TSED) at $T_\text{error}=2.0$px as $T_{2\text{px}}$, and average run time per scene as $\tau$.
The first two sections report numbers from \protect\cite{prabhu2023inpaint3d} and \protect\cite{liu2024infusion}, respectively, possibly using different evaluation code than SPIn-NeRF. The last section ensures direct comparability with consistent evaluation code;
see \S\ref{supp:sec:tsed-eval} for a comprehensive TSED analysis.
}
\label{tab:spinnerf}
\vspace{-18pt}
\end{table}

\begin{table*}[t]
\centering
\tablesize
\begin{tabular}{c|ccc|cc|cc|cc|c}
Method & PSNR $\uparrow$ & SSIM $\uparrow$ & LPIPS $\downarrow$ & MUSIQ $\uparrow$ & Corrs $\uparrow$ & $\sigma^D \uparrow$ & $\sigma^N \uparrow$ & $T^D_{2\text{px}} \uparrow$ & $T^N_{2\text{px}}\uparrow$ & $\tau \downarrow$ \\ \hline
Stable Diffusion (2D) \cite{stable.diffusion} &
             24.69 &              0.85 &              0.10 &              3.77 &
             1120 &              \rankonecolor44.55 &             25.55 & 9.81 & 86.55 & \rankonecolor3m \\
NeRFiller w/o depth \cite{weber2024nerfiller} &
             27.96 &              0.88 &              0.07 & 3.68              &
1146              & 1.20                  & 3.18               & 14.63             &
93.51              & 1h 30m \\
NeRFiller \cite{weber2024nerfiller}           &
27.68              & 0.87              & 0.08              &             3.69 &
             1185 & 1.25 & 3.31 & 16.53 & 96.04 & 1h 30m                      \\
Ours &
\rankonecolor28.59 & \rankonecolor0.89 & \rankonecolor0.05 & \rankonecolor3.80 &
\rankonecolor1250 & 38.96 & \rankonecolor26.45 & \rankonecolor67.80 &
\rankonecolor98.25 & 55m
\end{tabular}%
\vspace{-5pt}
\caption{Evaluation of scene completion on the NeRFiller dataset.
We denote sharpness ($\times 10^{-5}$) as $\sigma$, the percentage of consistent image pairs (TSED) at $T_\text{error}=2.0$px as $T_{2\text{px}}$, average run time per scene as $\tau$, and independent 2D inpainting as ``2D''. $\cdot^D$: direct outputs of the inpainting model - $\cdot^N$: NeRF renders; see \S\ref{supp:sec:tsed-eval} for a comprehensive TSED analysis.
}
\label{tab:nerfiller}
\vspace{-10pt}
\end{table*}

\begin{figure*}[t]
    \centering
    \includegraphics[width=\linewidth]{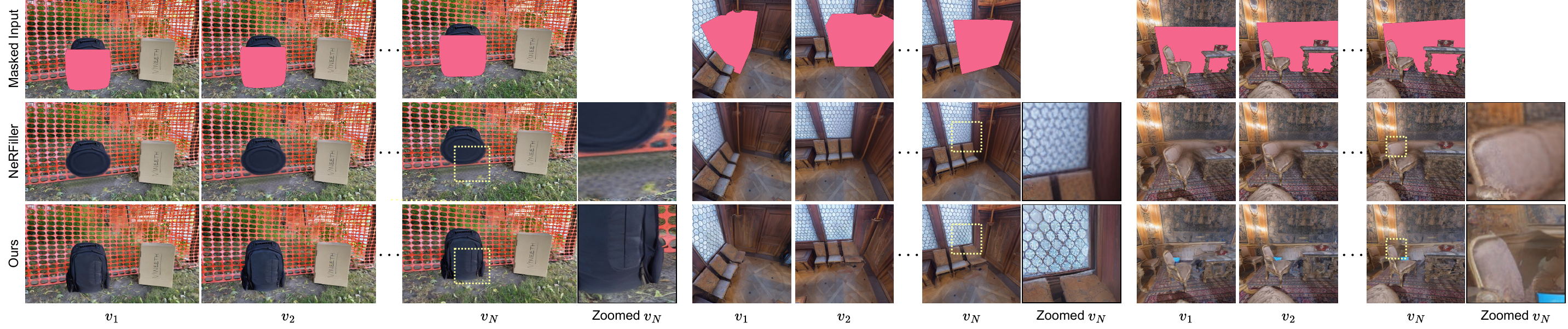}
    \vspace{-20pt}
    \caption{ 
    Qualitative scene completion comparisons on the NeRFiller dataset. 
    NeRFiller can converge to blurry content, due to mixing divergent views,
    while ours generates and then propagates sharp content 
    (e.g., see details in the backpack or window glass in the zoomed patches).
    Each view in a scene is denoted by $v_i$, where $i$ is the view index.
    }
        \vspace{-15pt}
    \label{fig:nerfiller-qualitative}
\end{figure*}

\noindent\textbf{Scene Completion with Wide Baselines.} 
The quantitative results for the scene completion task are presented in \cref{tab:nerfiller}, demonstrating the superiority of our method on all metrics. We also show that our method is less time-consuming than NeRFiller.
Moreover,
although fitting a NeRF 
on our inpaintings
reduces sharpness,
they
still
remain significantly sharper than
NeRFiller's.
We qualitatively illustrate this in \cref{fig:nerfiller-qualitative}.
To evaluate 3D consistency, we report TSED on two sets of images, (i) the NeRF datasets (i.e., direct outputs of the inpainting models), and (ii) the NeRF renders. For NeRFiller, we use the dataset produced in the latest Dataset Update iteration. As illustrated in 
\cref{tab:nerfiller},
as our dataset used to fit the NeRF is significantly more consistent than that of NeRFiller, the final NeRF achieves a higher consistency score.
We also significantly outperform the naive 2D-only baseline (independent inpaintings; see \S\ref{supp:subsec:independent-ablation} for details).

\begin{figure}[t]
    \centering
    \includegraphics[width=\linewidth]{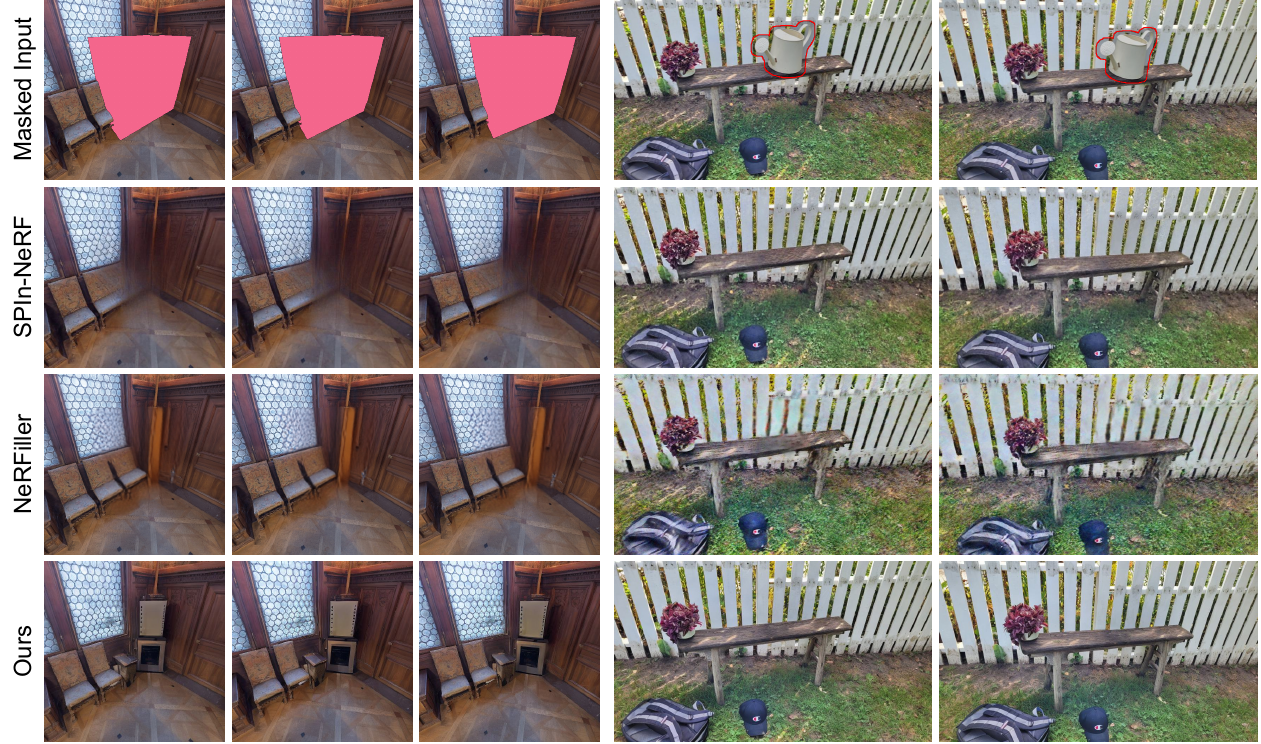}
    \vspace{-20pt}
    \caption{Qualitative comparisons for few-view inpainting. Our inpainted images are sharper and more visually plausible.
    }
    \vspace{-15pt}
    \label{fig:few-view-qualitative}
\end{figure}

\noindent\textbf{Inpainting with Few-view Inputs.}
We use single-reference inpainting for the few-view task. As depicted in \cref{tab:few-view}, we outperform all the baselines on the few-view inpainting task,
even without the need to use
the ground-truth camera parameters and depth maps,
making it more self-contained.
The TSED results demonstrate our method achieves a higher consistency score, mainly due to the other methods relying on fitting a NeRF, which is suboptimal for extremely sparse views. The qualitative results shown in \cref{fig:few-view-qualitative,fig:further-few-view} also confirm the higher quality of our inpainted images.

\begin{table}[t]
\centering
\tablesize
\setlength{\tabcolsep}{0.69em}
\begin{tabular}{ccc|ccc|c|c}
Method              & $\Pi$ & $D$ & $\sigma \uparrow$ & MUSIQ $\uparrow$ & Corrs $\uparrow$ & $T_{2\text{px}} \uparrow$ & $\tau \downarrow$ \\ \hline
SPIn-NeRF \cite{spinnerf}           & \Checkmark   & \XSolidBrush & 17.18         & 3.26                & 278 & 25.42                            & 15m      \\
NeRFiller \cite{weber2024nerfiller}          & \Checkmark   & \XSolidBrush & 5.05                         & 3.41                & 187                                            & 18.54 & 13m       \\
NeRFiller \cite{weber2024nerfiller}           & \Checkmark   & \Checkmark   & 5.41                         & 3.42 & 183                          & 18.96 & 13m       \\
Ours                & \XSolidBrush & \XSolidBrush & \rankonecolor48.2            & \rankonecolor3.84   & \rankonecolor 400 & \rankonecolor52.92 & \rankonecolor 12s     
\end{tabular}%
\vspace{-5pt}
\caption{Quantitative evaluation of the few-view task. We denote camera parameters as $\Pi$, depth maps as $D$, sharpness ($\times 10^{-5}$) as $\sigma$, the percentage of consistent image pairs (TSED) at $T_\text{error}=2.0$px as $T_{2\text{px}}$, and average run time per scene as $\tau$;
see \S\ref{supp:sec:tsed-eval} for a comprehensive TSED analysis.}
\label{tab:few-view}
\vspace{-10pt}
\end{table}

\subsection{Ablation Studies}
\label{subsec:ablation}

In \cref{tab:ablation-summary}, 
we summarize our extensive ablation studies from \S\ref{supp:sec:full-ablation}, ablating key components of our inpainting pipeline on scene completion.
We observe (first row) the importance of conditioning the inpainter on the geometric cues (\S\ref{subsubsec:cond-inp}).
For wide-baseline datasets like NeRFiller, our autoregressive procedure (\S\ref{subsec:autoregressive}) is essential, since a single reference lacks sufficient information for a wide baseline (second row).
Finally, we find that providing \duster with ground-truth depth maps has minimal impact; however, known camera parameters significantly improve performance (third and fourth rows). This is mainly because optimizing the camera parameters in \duster requires complex global alignment, while known cameras simplify depth optimization.

\begin{table}[t]
\centering
\tablesize
\setlength{\tabcolsep}{0.5em}
\begin{tabular}{c|ccc|cc}
Variation & PSNR $\uparrow$ & SSIM $\uparrow$ & LPIPS $\downarrow$ & MUSIQ $\uparrow$ & Corrs $\uparrow$ \\ \hline
No Geometric Cues   &              28.36 &              0.88 & \rankonecolor0.05 &              3.78 &              1223 \\
Single-Reference    &              27.42 &              0.88 &              0.07 &              3.77 &              1232 \\
No GT Camera        &              28.29 &              0.88 & \rankonecolor0.05 &              3.79 &              1231 \\
No GT Depth         & \ranktwocolor28.44 & \rankonecolor0.89 & \rankonecolor0.05 & \rankonecolor3.80 & \rankonecolor1252 \\
Full Model          & \rankonecolor28.59 & \rankonecolor0.89 & \rankonecolor0.05 & \rankonecolor3.80 & \ranktwocolor1250 \\
\end{tabular}
\vspace{-10pt}
\caption{
Summary of key ablations for various design choices in the model and inpainting procedure on the NeRFiller dataset.
Please see \S\ref{subsec:ablation} for an explanation and \S\ref{supp:sec:full-ablation} for our exhaustive ablation studies.}
\label{tab:ablation-summary}
\vspace{-15pt}
\end{table}

\section{Conclusion}

In this paper, we introduced a novel approach to 3D scene inpainting task, \textit{without}
a 3D radiance field.
While this avoids fusing in pixel space, which induces blurriness, it necessitates a novel way to fuse information across views via estimated scene geometry.
We do so via training a diffusion-based inpainter, conditioned on appearance and geometric cues from a reference view-set, which enables fusion in the learned space of the generative model, thus retaining sharpness.
The resulting multiview inpainting algorithm is highly versatile, 
capable of handling the sparse-view inpainting task, on which other methods struggle, and able to operate without camera poses.
We explored the efficacy of our method on two recent benchmarks, encompassing narrow- and wide-baseline 3D scenes, as well as the few-view scenario, showing state-of-the-art performance in all cases, in terms of both image quality and multiview consistency.

\section*{Acknowledgements}

This work 
is supported by
the Vector Scholarship in Artificial Intelligence (A.S.) provided through the Vector Institute,
the Canada First Research Excellence Fund (CFREF) for Vision: Science to Applications (VISTA) program (A.S., T.A.A., M.A.B., and K.G.D.),
and the NSERC Discovery Grant program (M.A.B., and K.G.D.).

{
    \small
    \bibliographystyle{ieeenat_fullname}
    \bibliography{defs,main}
}

\newpage\clearpage
\setcounter{section}{0}
\renewcommand{\thesection}{\Alph{section}}
\renewcommand{\theHsection}{\Alph{section}}
\maketitlesupplementaryarxiv

\section{Overview}

This document provides additional supplemental material to the main paper.
\S\ref{supp:sec:method} details our proposed method, including scene geometry estimation, triangle mesh creation, the fusion of multiple reference images, training details, and inference details.
\S\ref{supp:sec:impl} outlines additional implementation details for our approach as well as the hyperparameters used. \S\ref{supp:sec:dataset} and \S\ref{supp:sec:metrics} describe details on the datasets and metrics used for evaluation, respectively.
\S\ref{supp:sec:qualitative} provides additional qualitative results.
\S\ref{supp:sec:tsed-eval} provides comprehensive results using the TSED metric.
\S\ref{supp:sec:full-ablation} demonstrates comprehensive ablation studies on various design decisions for our inpainting pipeline.
Finally, \S\ref{supp:sec:limitations} discusses the limitations of our work.
For more qualitative demonstrations,
please see our project page: \href{https://geomvi.github.io/}{https://geomvi.github.io}.

\section{Methodological Details}
\label{supp:sec:method}

\subsection{\duster as an Autoregressive Scene Geometry Estimator}
\label{supp:subsec:dust3r}

\duster \cite{dust3r} uses a network to predict 3D pointmaps for image pairs, followed by a global optimization to obtain the global camera parameters and dense depth maps. \duster is not trained on incomplete views (i.e., not-yet-inpainted images); however, we adapt it to use incomplete views by passing the images to the model with the masked area set to zero. Then, we suppress DUSt3R's predicted confidence maps in the masked area by setting the confidence to zero. This will prevent incomplete parts from contributing during the optimization phase.

Optionally, we also pre-set ground-truth information for the optimization phase. For pre-setting camera parameters, we initialize and freeze the corresponding parameters in the optimization. For pre-setting incomplete dense depth, we only initialize the depth (wherever provided), while allowing it to change during the optimization.

For small-baseline scenes and the few-view task, we compute \duster on a complete symmetric scene graph $G = (V, E)$, connecting each view to all other views. However, for large-baseline scenes, we restrict the connections of each view to their $k$ closest views. We use the rotation (orientation) difference between the cameras as our camera distance measure, as the cameras nearly always point to a central point in the scene.
Specifically, for an edge $e = (i, j)$ with corresponding extrinsic rotation matrices $\RB_i, \RB_j \in \SO(3)$, we define the view distance function as
\begin{equation}
d(e) = \left\| A\left(\RB_i^\top \RB_j\right) \right\|_2,
\label{eq:view-distance}
\end{equation}
where $A: \SO(3) \to [0, 2\pi) \times [0, 2\pi) \times [0, \pi)$ is a function mapping a rotation matrix to Euler angles. In other situations, this heuristic could be altered (e.g., to use an estimate of overlapping image content, or camera positional information). Therefore, for wide-baseline scenes, the set of edges is
\begin{equation}
E = \{(i, j) \in V^2 ; i \neq j, j \in \topk_j (V, -d(i, j)) \},
\end{equation}
where $\topk_i (S, f(i))$ denotes the top $k$ elements of the set $S$ according to function $f$.

In the first autoregressive step, before any image is inpainted, we run \duster on the entire scene graph. This will yield the optimized camera parameters, $(\RB_i, \tB_i) \in \SE(3), \KB_i \in \real^{3\times 3}$, dense depth maps, $D_i \in \real^{H \times W}$, and DUSt3R's pairwise poses, $(\RB_e, \tB_e) \in \SE(3)$. We denote the set of optimized geometry parameters as 
\begin{equation}
\mathcal{G} = \{\RB_i, \tB_i, \KB_i, D_i\}_i \cup \{\RB_e, \tB_e\}_e.
\end{equation}

In each remaining step, where the views $V_I \subset V$ are being inpainted, we only update their corresponding views, $V_I$, and edges, $E_I = \{(i, j) \in E ; i \in V_I \}$, of the scene graph, initialized by the current state of $\mathcal{G}$, while freezing other parameters. We then replace the updated parameters in $\mathcal{G}$. This greatly reduces the run-time of the geometry update, which is significant, since it is performed after every iteration of autoregressive inpainting.

\subsection{Creating Triangle Meshes from Dense Depth Maps}
\label{supp:subsec:mesh}

Given the camera parameters $(\RB, \tB) \in \SE(3), \KB \in \real^{3\times 3}$, and dense depth map $D \in \real^{H\times W}$, we can lift the pixel coordinates to the world coordinate system via the lifting function
\begin{align}
\XB(\xB) &:= \RB^\top \lp
    D[\xB] \KB^{-1} h(\xB) - \tB
\rp,
\end{align}
where $h: \real^n \to \real^{n + 1}$ is the homogeneous mapping. Following GeoFill \cite{zhao2023geofill}, we build a triangle mesh with a regular grid, where the mesh vertices are provided by the lifting function, $\XB(\xB)$. To create discontinuities on object boundaries, we drop the mesh edges wherever there is a sudden change in depth. We use the same criterion as GeoFill \cite{zhao2023geofill}, where we drop the edge between two adjacent vertices $v_i, v_j$ if
\begin{align}
\frac{
    2 \left|
        D[\xB_i] - D[\xB_j]
    \right|
}{
    D[\xB_i] + D[\xB_j]
} > \epsilon_\text{edge},
\label{eq:depth-criterion}
\end{align}
where $\epsilon_\text{edge} = 4 \times 10^{-2}$ is a user-defined hyperparameter. This results in the 3D mesh, $\mesh$.

\subsection{Creating a Shadow Volume Mesh using Silhouette Edges}
\label{supp:subsec:shadow}

\begin{figure}[t]
    \centering
    \includegraphics[width=\linewidth]{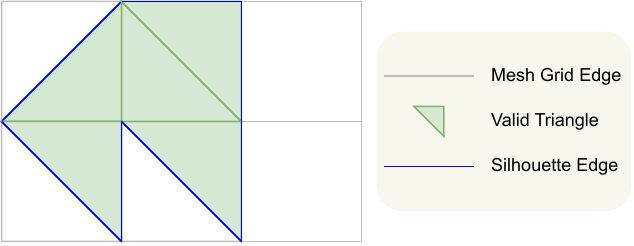}
    \caption{Illustration of the silhouette edges in a triangle mesh. An edge that only belongs to \emph{one} triangle is considered a silhouette edge.}
    \label{fig:silhouete-edges}
\end{figure}

To create the shadow volume mesh for a reference view, we first identify the silhouette edges of the mesh. When building the triangle mesh as described in \S\ref{supp:subsec:mesh}, we also identify the triangle edges at surface boundaries (i.e., silhouette edges) by looking for edges that only belong to one triangle. \cref{fig:silhouete-edges} illustrates the selection of silhouette edges. Let $E_\mathcal{S} \subset \real^{2 \times 3}, \left|E_\mathcal{S}\right| < \infty$ denote the set of silhouette edges in the world coordinate system. Given the camera extrinsics, the camera center is at $-\RB^\top \tB$ in the world coordinate system. Our goal is to draw a ray from the camera center to each point on the edge of the shape (i.e., its occlusion boundaries), and then extend the ray past it (forming part of the boundary of the shadow volume induced by that geometric element; see Fig.~\ref{fig:autoregressive} for a visualization).
For each vertex, $v$, in a silhouette edge, we extrude it along its corresponding ray direction according to the reference camera, as
\begin{equation}
v' = v + \varepsilon_d \frac{
    v + \RB^\top \tB
}{
    \left\|
        v + \RB^\top \tB
    \right\|_2
},
\end{equation}
where $\varepsilon_d$ 
is a sufficiently large value to ensure the rendered shadow volume covers the relevant part of other views. For each silhouette edge, $(v_1, v_2) \in E_\mathcal{S}$, we form a quad, $(v_1, v_2, v'_2, v'_1)$, forming the side walls of the shadow volume. We then split each shadow quad into two triangles and use the resulting triangle mesh to render the shadow volumes.

\subsection{Details on the Hint Image}
\label{supp:subsec:hint}

As mentioned in \S\ref{subsubsec:cond-inp}, we optionally use a hint image, $H$, to provide style information to the network. For inpainting the first image, as we do not have any inpainted images, we use an empty image (zero-valued) as the hint. For all other steps, we select the furthest inpainted image to the inpainting image $i$ as
\begin{equation}
H = \argmax_{h \in \mathcal{I}} d(i, h),
\end{equation}
where $\mathcal{I}$ is the set of inpainted images, and $d$ is the view distance function defined in \cref{eq:view-distance}.

\subsection{Parallel Process Fusion via Predicted Confidence Maps}
\label{supp:subsec:fusion}

\begin{figure*}[t]
    \centering
    \includegraphics[width=\linewidth]{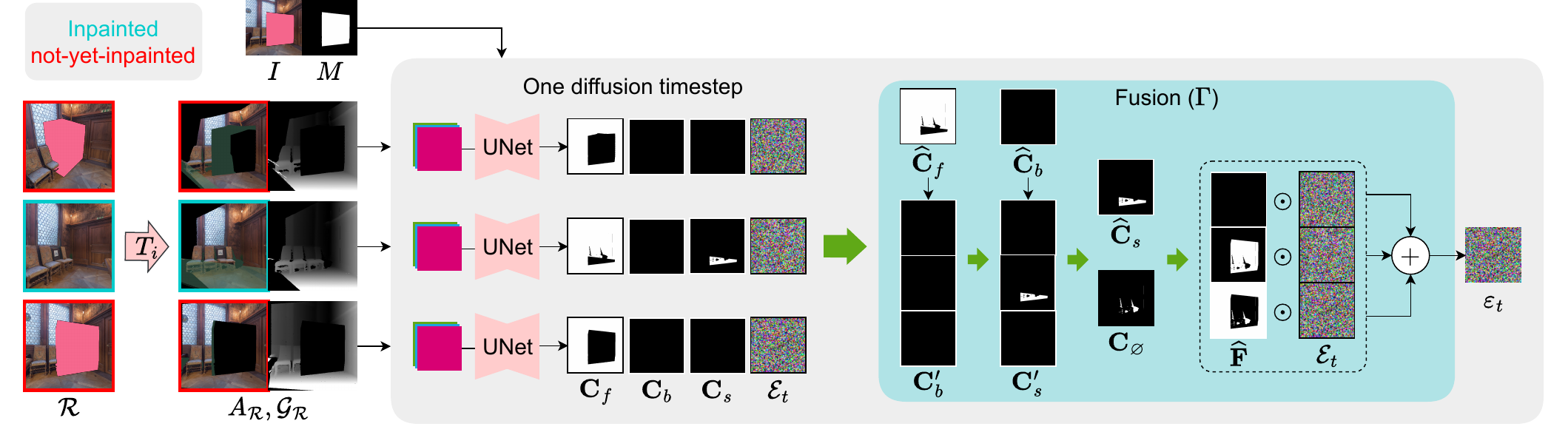}
    \caption{Illustration of the fusion of multiple reference views using the confidence masks. We denote $I$ as the \emph{incomplete} target image, $M$ as the inpainting mask, $\mathcal{R}$ as the set of reference images, $(A_\mathcal{R}, \mathcal{G}_\mathcal{R})$ as reference-based appearance and geometric cues, $\Gamma$ as the fusion operator, $\CB_f, \CB_b, \CB_s$ as the front-face, back-face, and shadow confidence mask, $\mathcal{E}_t$ as the noise estimates at diffusion timestep $t$, $\widehat{\FB}$ as the fused mask, and $\varepsilon_t$ as the fused noise estimate at diffusion timestep $t$, respectively. $\widehat{\CB}_f, \widehat{\CB}_b, \widehat{\CB}_s, \CB'_b, \CB'_s, \CB_\varnothing$ represent intermediate variables of the fusion process; see \cref{alg:fusion} for details. As shown, given a target image and a set of reference images, we first render the reference-based appearance and geometric cues, and then at each diffusion step, we fuse the noise estimates conditioned on different reference views using the predicted confidence masks.}
    \label{fig:fusion}
\end{figure*}

\cref{alg:fusion} summarizes the steps taken by the fusion operator, $\Gamma$. When inpainting a target view, $\tau$, conditioned on a set of reference views, $\mathcal{R}$, in addition to the noise estimates, $\mathcal{E}_t$, and their corresponding confidence masks, $\mathcal{C}_t$, we also utilize the view distances of the reference views to the target view, computed as
\begin{equation}
d_r = \left\{
    d(\tau, r)
\right\}_{r \in \mathcal{R}},
\end{equation}
where $d$ is the view distance function defined in \cref{eq:view-distance}. 
\cref{fig:fusion} visualizes the fusion operator.

\begin{algorithm*}
\caption{Pseudo-code for fusing the noise estimates, each conditioned on a specific reference view. We denote $\mathcal{E}_t$ as the noise estimates, each conditioned on a specific reference view at diffusion timestep $t$, $\CB_f, \CB_b, \CB_s$ as the front-facing, back-facing, and shadow confidence masks, $d_r$ as the view distances of the reference views to the target view, $R$ as the number of reference images used to inpaint the image, $\lor, \land, \neg$ as logical ``or'', ``and'', and ``negation'', $\odot, \oslash$ as Hadamard product and division, and $\onehot(i, N): \integer^{\cdots} \to \binary^{N \times \cdots}$ as a function that encodes an index $i$ into an $N$-length one-hot vector, respectively. For the shadow confidence mask, ``one'' means that although the ray intersects the shadow volume, and the content is \textit{uncertain}, the model has decided that there is no occluded content, and the shadow background is valid; see \S\ref{supp:subsec:training}. }
\begin{algorithmic}[1]
\Procedure{$\Gamma$}{$\mathcal{E}_t \in \real^{R \times C \times H \times W},
\left\{ \CB_f, \CB_b, \CB_s \right\} \subset \binary^{R \times H \times W}, d_r \in \real^R$}
    \State $\widehat{\CB}_f = \bigvee_r \CB_f[r, :, :]$
    \Comment{At least one front-face exists. $\in \binary^{H \times W}$}
    \State $\CB'_b = \CB_b \, \land \, \neg \widehat{\CB}_f$
    \Comment{Back-faces but not front-faces. $\in \binary^{R \times H \times W}$}
    \State $\widehat{\CB}_b = \bigvee_r \CB'_b[r, :, :]$
    \Comment{At least one back-face exists. $\in \binary^{H \times W}$}
    \State $\CB'_s = \CB_s \, \land \, \neg (\widehat{\CB}_f \lor \widehat{\CB}_b)$
    \Comment{Shadows but no front-faces or back-faces. $\in \binary^{R \times H \times W}$}
    \State $\widehat{\CB}_s = \bigvee_r \CB'_s[r, :, :]$
    \Comment{At least one shadow exists. $\in \binary^{H \times W}$}
    \State $\CB_\varnothing = \neg (\widehat{\CB}_f \lor \widehat{\CB}_b \lor \widehat{\CB}_s)$
    \Comment{No confidence. $\in \binary^{H \times W}$}
    \State $\CB = \concat(\CB_f, \CB'_b, \CB'_s, \CB_\varnothing)$
    \Comment{Confidence hierarchy. $\in \binary^{4 \times R \times H \times W}$}
    \State $\WB = \CB \oslash d_r$
    \Comment{Weight map (prefer closer views). $\in \real^{4 \times R \times H \times W}$}
    \State $\FB = \sum_{i=1}^4 \left(
        \argmax_r \WB[i, r, :, :]
    \right) \odot \left(
        \bigvee_r \CB[i, r, :, :]
    \right)$
    \Comment{Selected reference indices. $\in \integer^{H \times W}$}
    \State $\widehat{\FB} = \onehot(\FB, R)$
    \Comment{Fused mask. $\in \binary^{R \times H \times W}$}
    \State $\varepsilon_t = \sum_r \mathcal{E}_t[r, :, :, :] \odot \widehat{\FB}[r, :, :]$
    \Comment{Fused noise estimate. $\in \real^{C \times H \times W}$}
    \State \Return $\varepsilon_t$
\EndProcedure
\end{algorithmic}
\label{alg:fusion}
\end{algorithm*}

\subsection{Training Details}
\label{supp:subsec:training}

\noindent\textbf{Single-View Data Synthesis.}
As mentioned in \S\ref{subsec:inptrain}, we synthesize the geometric and appearance cues from single-view images, via monocular depth estimation. Specifically, given an image, $I \in \real^{3 \times H \times W}$, we first compute a monocular metric depth estimate, $D \in \real^{H \times W}$. We assume the focal length to be $f = \frac{W + H}{2}$, and the principal point to be in the center of the image, $\pB = \frac{1}{2} (W, H)$. We then create a triangle mesh, $\mesh$, in the image's coordinate frame (i.e., identity extrinsics), as described in \cref{supp:subsec:mesh}. We assume there is a second camera (i.e., a synthetic reference view) from which the geometric and photometric cues are actually coming. To generate a random reference pose, we sample angles, $\aB_r \sim \mathcal{N}\left(0, \sigma_{a_r}^2 \eye_3\right)$, and translation, $\tB_r \sim \mathcal{N}\left(0, \left(\sigma_{t_r} \cdot \min D\right)^2 \eye_3\right)$, where $\sigma_{a_r} = 0.3$ and $\sigma_{t_r} = 0.2$ are user-defined hyperparameters. We then form the rotation matrix, $\RB_r$, from the sampled Euler angles, $\aB_r$, and render the mesh, $\mesh$, to obtain the synthetic reference image, $I_r$, and its corresponding depth map, $D_r$. This process will automatically occlude parts of the target view. We do not use hint images when the training sample is from a single-image dataset.

\noindent\textbf{3D Data Sampling.}
For a mesh-based 3D dataset, given a mesh, $\mesh$, we uniformly sample the reference and target cameras (in a sphere centered at the object's centroid), directly rendering the reference and target views, $I_r, I$, respectively, along with the reference depth map, $D_r$. With a probability of $p_h = 0.95$, we also uniformly sample another camera to render a hint image for training.

\noindent\textbf{Mask Generation.}
We consider two mask generation strategies:
(i) the 2D image-based approach from LaMa \cite{lama},
which generates large and diverse masks on the target image,
and
(ii) a 3D-based approach, designed to obtain a 3D consistent mask across a multiview image set.
We focus on (ii) for the remainder of this section.
This strategy is straightforward, given known 3D geometry:
we sample a 3D convex polyhedron, place it in the scene, and obtain an inpainting mask by rendering this occluder volume to the target view.
Specifically, let $B \in \real^{2 \times 3}, C \in \real^3$ be the bounding box around the scene point cloud and the scene point cloud's centroid, respectively. We first uniformly sample a bounding box, $B_o$, for the occluder inside the scene's bounding box. We restrict the size of the bounding box to be in the range
\begin{equation}
\left[
    o_\text{min} (B_2 - B_1), o_\text{max} (B_2 - B_1)
\right],
\end{equation}
where $o_\text{min}$ and $o_\text{max}$ are user-defined hyperparameters. For Google Scanned Object \cite{gso.dataset}, we set $o_\text{min} = 0.6$ and $o_\text{max} = 1.0$; however, as MS COCO \cite{coco.dataset} includes outdoor scenes, the scene's bounding box does not properly represent the scene's boundaries. In that case, we first uniformly sample a point from the scene's point cloud as the occluder's centroid, $C_o$. We also restrict the sampled occluder bounding box to the camera frustum, instead of the scene's bounding box. To that end, we restrict the size of the occluder bounding box to be in the range
\begin{equation}
\left[
    o_\text{min} \frac{\left(C_o^\top \uvec{k}\right) (B_2 - B_1)}{B_2 - B_1}, o_\text{max} \frac{\left(C_o^\top \uvec{k}\right) (B_2 - B_1)}{B_2 - B_1}
\right],
\end{equation}
where $\uvec{k}$ is the unit vector in the direction of the $z$-axis, and we set $o_\text{min} = 0.6$ and $ o_\text{max} = 0.8$. We then uniformly sample $N_o$ points within $B_o$, and fit a convex hull around the sampled points, resulting in the occluder volume. We render the sampled convex hull to the target view, yielding the inpainting mask. With a probability of 0.2, we sample a 3D occluder volume; otherwise, we use LaMa's mask generator.

\noindent\textbf{Simulating Geometric Errors for Domain Adaptation.}
Our data generation techniques, whether on 3D scenes or single images, are slightly out-of-distribution compared to real-world multiview datasets (on which our method is evaluated).
In particular, the geometric errors (whether in scene or camera parameters) from \duster are not naturally present.
We therefore consider how to include such errors synthetically.

Given a reference image, $I_r$, and its corresponding depth map, $D_r$, we create a triangle mesh, $\mesh_r$, along with its shadow mesh, $\mathcal{S}_r$, (as described in \S\ref{supp:subsec:mesh} and \S\ref{supp:subsec:shadow}). To simulate geometry estimation errors, we also create a perturbed version of the reference mesh, $\mesh'_r$, and shadow mesh, $\mathcal{S}'_r$, by sampling perturbation angles, $\aB_p \sim \mathcal{N}\left(0, \sigma_{a_p}^2 \eye_3\right)$, and translation, $\tB_p \sim \mathcal{N}\left(0, \left(\sigma_{t_p} \cdot \min D_r\right)^2 \eye_3\right)$, forming the rotation matrix, $\RB_p$, and perturbing the mesh vertices, $\VB_r \in \real^{V \times 3}$, as
\begin{equation}
\VB'_r = \VB_r \RB_p^\top + \tB_p,
\end{equation}
where $\sigma_{a_p} = 0.2$ and $\sigma_{t_p} = 0.01$ are user-defined hyperparameters. We then render the perturbed mesh and shadow to obtain the rendered appearance cue, $T_r(I_r)$, and geometric cues, $\mathcal{G}_\mathcal{R} = \{F_r, B_r, \widehat{D}_r, C_r\}$. 

\noindent\textbf{Supervising Predicted Confidence Masks.}
We use the unperturbed meshes, $\mesh_r$, and $\mathcal{S}_r$, to render the ground-truth confidence masks. Specifically, we render $\mesh_r$ and $\mathcal{S}_r$ to obtain the unperturbed front-face, back-face, and shadow masks, $\widehat{F}_r, \widehat{B}_r, \widehat{C}_r$. Given the sampled inpainting mask, $M$, the ground-truth front-face confidence mask is computed as
\begin{equation}
\CB_f = (\widehat{F}_r \land \neg \widehat{C}_r) \lor \neg M,
\end{equation}
where $\land, \lor, \neg$ denote the logical ``and'', ``or'', and negation, respectively. This mask highlights the regions that are either outside the inpainting mask, or exclusively guided by front-facing surfaces (excluding the parts intersecting the shadow mask). The ground-truth back-face confidence mask is computed as
\begin{equation}
\CB_b = \widehat{B}_r \land M.
\end{equation}
This mask highlights the regions inside the inpainting mask that are guided by back-facing surfaces.
To compute the ground-truth shadow confidence mask, first note that this mask indicates the model's certainty in trusting the photometric information. To obtain such information, let $\mathcal{F}, \mathcal{F}_r \in \integer^{H \times W}$ be the map of triangle face indices rendered to the target view, from meshes $\mesh$ and $\mesh_r$, respectively. We can trust the photometric information of a pixel, if and only if $\mathcal{F}$ and $\mathcal{F}_r$ are equal in that pixel and the pixel has valid photometric information (i.e., not disoccluded), meaning both reference and target see the same triangle face at that pixel. Therefore, the ground-truth shadow confidence mask is computed as
\begin{equation}
\CB_s = \widehat{C}_r \land \mathds{1}\left\{\mathcal{F}[\xB] = \mathcal{F}_r[\xB]\right\}_\xB \land \widehat{F}_r \land M,
\end{equation}
where $\mathds{1}\{\cdot\}$ denotes the indicator function.

\noindent\textbf{Complete Loss Function.}
After obtaining all input and corresponding ground-truth outputs, we sample a random noise, $\varepsilon$, and timestep, $t$, training the model via %
\begin{align}
&I_M = I \odot \neg M, \\
&\{ \widetilde{\varepsilon}, \widetilde{\CB}_f, 
    \widetilde{\CB}_b, \widetilde{\CB}_s \} = 
    \epsilon_\theta\left(
        z_t, M, I_M, A_\mathcal{R}, \mathcal{G}_\mathcal{R}, y, t
    \right), \\
&\mathcal{L}(\theta) =
\left\|
    \varepsilon - \widetilde{\varepsilon} 
\right\|_2^2 
+
\sum_{\rho\in \{ f,b,s \} }
\left\|
    \CB_\rho - \widetilde{\CB}_\rho
\right\|_2^2,
\end{align}
where $ \mathcal{L} $ is the training loss, $I_M$ is the masked input, $z_t = \addNoise(I, \varepsilon, t)$ is the forward diffusion step, yielding the latent noisy diffusion intermediate, and $A_\mathcal{R}, \mathcal{G}_\mathcal{R}, y$ denote the appearance cues, geometric cues, and the text prompt, respectively. Text prompts, $y$, are already provided as captions in MS COCO \cite{coco.dataset}, and we generate them for GSO \cite{gso.dataset}, as mentioned in \S\ref{supp:sec:impl}.

\begin{algorithm*}
\caption{Pseudo-code for selecting a wide-baseline subset of the scene. $\mathcal{D}, i_1, \setminus$ denote the view distance matrix, the initial image to be inpainted, and set subtraction, respectively.}
\begin{algorithmic}[1]
\Procedure{SelectWideBaseline}{$\mathcal{D} \in \real^{N \times N}, i_1 \in \integer$}
    \State $W = [i_1]$
    \Comment{Initialize the wide-baseline set}
    \For{$n$ in $[1, \cdots, N-1]$}
        \State $i_n = \argmax_{i \notin W} \min_{j \in W} \mathcal{D}[i, j]$
        \Comment{Find the next wide-baseline view using min-max}
        \State $W = \concat(W, [i_n])$
        \Comment{Extend the wide-baseline set}
    \EndFor
    \\
    \State $\widehat{W} = [i_1]$
    \Comment{Initialize the sorted wide-baseline set}
    \For{$n$ in $[1, \cdots, N-1]$}
        \State $i_n = \argmin_{i \in W \setminus \widehat{W}} \frac{1}{|\widehat{W}|} \sum_{j \in \widehat{W}} \mathcal{D}[i, j]$
        \Comment{Minimize the mean distance to the previous sorted views}
        \State $\widehat{W} = \concat(\widehat{W}, [i_n])$
        \Comment{Extend the sorted wide-baseline set}
    \EndFor
    \State \Return $\widehat{W}$
\EndProcedure
\end{algorithmic}
\label{alg:wide-baseline}
\end{algorithm*}

\subsection{Selecting a Wide-baseline Subset}
\label{supp:subsec:wide-baseline}

As mentioned in \S\ref{subsubsec:next-images}, we initially inpaint a wide-baseline subset of the scene, one by one. Let $N$ be the number of views in the scene. We first form the view distance matrix
\begin{equation}
\mathcal{D} = \begin{bmatrix}
d(i, j)
\end{bmatrix}_{i, j = 1}^{N},
\end{equation}
where $d$ is the view distance function defined in \cref{eq:view-distance}. We also randomly select an image, $i_1$, to start the inpainting from. As summarized in \cref{alg:wide-baseline}, we perform a greedy min-max approach to select the wide-baseline subset, and then sort the selected subset such that the view in each iteration, minimizes the mean distance to the views in the previous steps.

\section{Implementation Details}
\label{supp:sec:impl}

As mentioned in \S\ref{subsec:inptrain}, we initialize our reference-guided inpainter as a Stable Diffusion v2, fine-tuned for inpainting \cite{stable.diffusion,sdinp}.
In training, we upweight Google Scanned Objects (GSO) so that the ratio of GSO to MS COCO samples in each epoch is 1/10. Since GSO lacks text captions, we use Kosmos-2 \cite{kosmos-2} to caption a front-facing view of each object. To make the model robust towards color and texture discrepancies across multiple views, we randomly augment the texture of the reference mesh using color jitter with a factor of 0.1 for brightness, contrast, saturation, and hue. To preserve classifier-free guidance \cite{ho2022classifier} capabilities, we randomly drop the text prompt with a probability of 0.1. If the generated mask for an image is a 3D occluder volume, we randomly drop the \textit{reference} content occluded by the occluder volume with a probability of 0.2. This will enable conditioning the model on not-yet-inpainted reference views. Notice that our model must be capable of both transferring reference information and also inpainting when no reference information is present.
We use AdamW \cite{adamw} with a learning rate of $10^{-4}$, weight decay of $0.01$, $\beta_1 = 0.9$, and $\beta_2 = 0.999$. 
We train for approximately 8000 iterations, with a batch size of 96, and two gradient accumulation steps, on 16 NVIDIA L40 GPUs. 
The weight of the loss for the latent patches \textit{out}side the inpainting mask is set to $0.1$. During inference, we adopt the DDIM sampler \cite{ddim} with 50 denoising steps.

\section{Dataset Details}
\label{supp:sec:dataset}

\paragraph{Scene Completion with Wide Baselines.}
As mentioned in \S\ref{subsec:eval-protocol}, we use the \textit{scene-centric} portion of the NeRFiller dataset for the wide-baseline scene-completion task. Specifically, we use the following scenes:

\begin{itemize}
    \item ``backpack''
    \item ``billiards''
    \item ``drawing''
    \item ``norway''
    \item ``office''
\end{itemize}

\section{Metrics Details}
\label{supp:sec:metrics}

\paragraph{TSED.}
TSED evaluates the consistency of adjacent pairs in a view set \cite{yu2023long}. On the SPIn-NeRF dataset, as the scenes have very small baselines, all possible view pairs are considered adjacent views. Unlike \cite{yu2023long}, which considers a minimum of 10 feature matches for consistency, we only consider a minimum of two feature matches, as the inpainting mask is significantly smaller than the whole frame.

\section{Further Qualitative Results}
\label{supp:sec:qualitative}

\begin{figure*}[t]
    \centering
    \includegraphics[width=\linewidth]{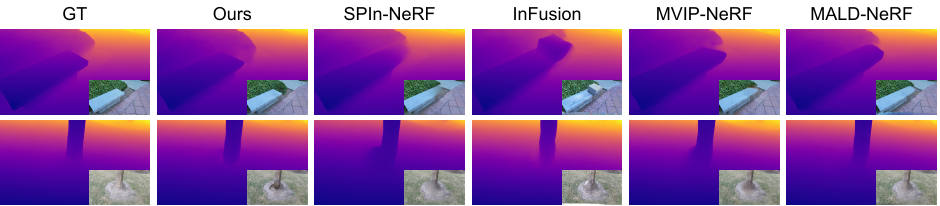}
    \caption{Visualized depth maps on SPIn-NeRF dataset. The depth maps are obtained by running \duster on the inpainted images. The corresponding inpainted images are also shown.}
    \label{fig:depth-vis}
\end{figure*}

\begin{figure*}
    \centering
    \includegraphics[width=\linewidth]{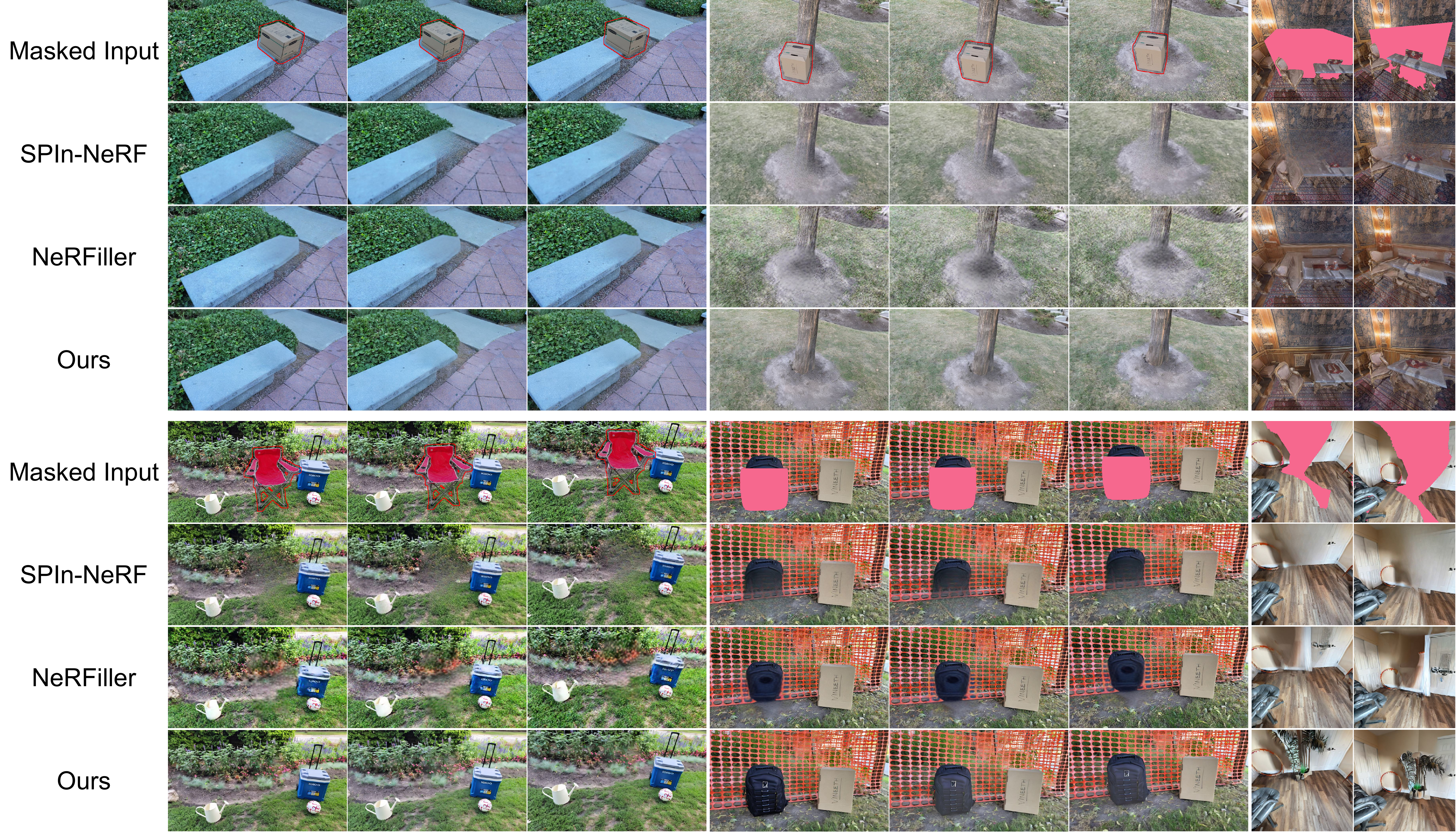}
    \caption{Further qualitative results for the few-view inpainting task. Please zoom in for details.}
    \label{fig:further-few-view}
\end{figure*}

In this section, we provide additional qualitative results.

\subsection{Comparison of Depth Maps}
\cref{fig:depth-vis} visualizes examples of inpainted images' depth maps from different inpainting methods. As shown, our method yields realistic geometry.

\subsection{Few-View Inpainting}
\cref{fig:further-few-view} presents further qualitative results for the few-view inpainting task, confirming higher sharpness and visual plausibility than the baselines.

\subsection{An Example of Autoregressive Inpainting}
\label{supp:subsec:autoregressive-qualitative}
\begin{figure*}[t]
    \centering
    \includegraphics[width=\linewidth]{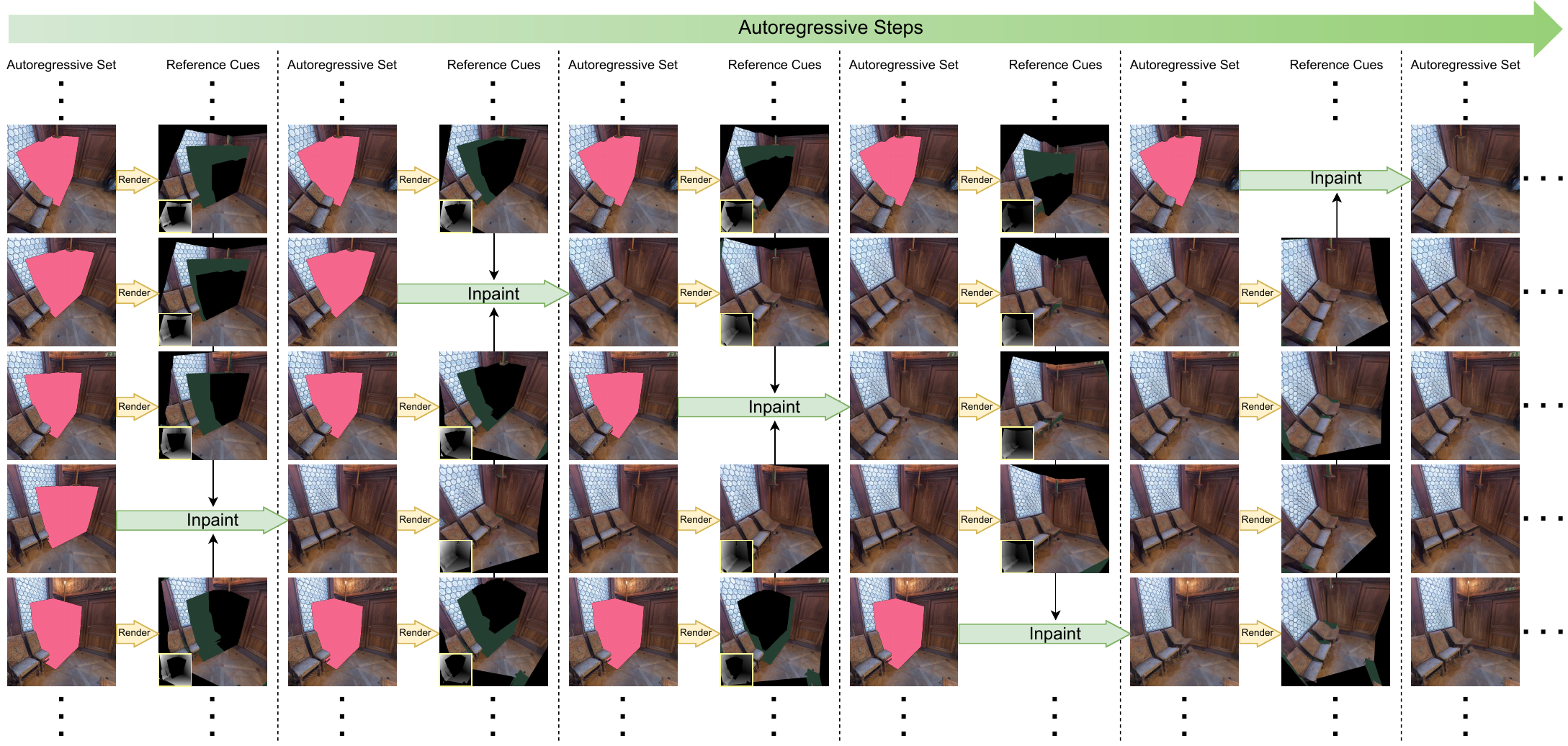}
    \caption{
    A step-by-step illustration of autoregressive inpainting in the first stage, where a wide-baseline subset of the scene is progressively inpainted. Note the consistency preserved throughout the process.
    }
    \label{fig:autoregressive-qualitative}
\end{figure*}

\cref{fig:autoregressive-qualitative} shows a step-by-step example of autoregressive inpainting progress in the first autoregressive stage, inpainting a wide-baseline subset of the scene. As detailed in \S\ref{subsubsec:next-images}, we begin with a random view and select a wide-baseline subset of the scene, progressively inpainting the closest view from this subset to the already inpainted ones at each autoregressive step.

\subsection{Qualitative Comparison with MALD-NeRF}
\label{supp:subsec:maldnerf-artifacts}
\begin{figure}[t]
    \centering
    \includegraphics[width=\linewidth]{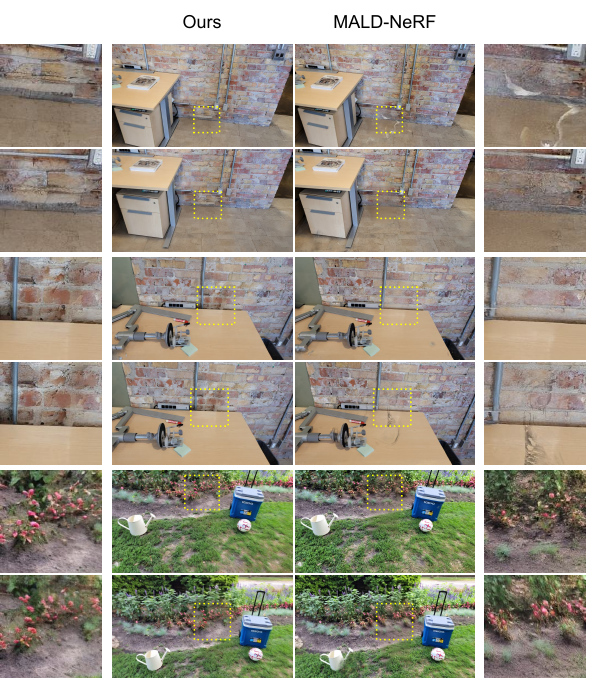}
    \caption{
    Qualitative comparison of our method with MALD-NeRF. Each inpainted view is accompanied by a zoomed-in version on the sides, highlighting inconsistencies.
    }
    \label{fig:maldnerf-artifacts}
\end{figure}

\cref{fig:maldnerf-artifacts} presents a direct qualitative comparison to MALD-NeRF \cite{lin2025taming}, the current state of the art. 
Here, we highlight the inconsistencies in MALD-NeRF's outputs, primarily caused by common NeRF artifacts, such as floaters or flawed geometry, which compromise geometric plausibility. As a result, the artifacts in MALD-NeRF prevent SIFT from detecting sufficient high-quality feature correspondences, leading to a lower TSED metric.

\section{Comprehensive TSED Evaluation}
\label{supp:sec:tsed-eval}

\begin{figure}[t]
    \centering
    \begin{subfigure}{\linewidth}
        \pgfplotsset{width=0.85\linewidth,height=4cm,compat=1.18}
\begin{tikzpicture}
    \begin{axis}[
        title style={align=center, font=\scriptsize, yshift=-.5em},
        title={TSED $\uparrow$},
        xlabel={$T_\text{error}$ (pixels)},
        ylabel={Consistency \%},
        xmin=0.9, xmax=4.1,
        ymin=30, ymax=80,
        xtick={1,2,3,4},
        ytick={35,45,55,65,75},
        legend pos=outer north east,
        legend style={nodes={scale=0.5, transform shape}},
        label style={font=\scriptsize},
        tick label style={font=\tiny},
        ymajorgrids=true,
        grid style=dashed,
        xlabel style={yshift=1ex},
        ylabel style={yshift=-1.5ex},
        mark size=1.5pt,x
    ]
        \addplot[color=blue,mark=*,] coordinates {
        (1.0,55.6154)(2.0,67.3590)(3.0,74.1795)(4.0,78.5801)
        };
        \addplot[color=red,mark=square*,] coordinates {
        (1.0,56.6538)(2.0,61.0449)(3.0,64.0769)(4.0,66.6731)
        };
        \addplot[color=orange,mark=pentagon*,] coordinates {
        (1.0,31.0385)(2.0,35.8846)(3.0,39.3974)(4.0,42.3558)
        };
        \addplot[color=violet,mark=oplus,] coordinates {
        (1.0,53.2949)(2.0,58.3269)(3.0,61.7564)(4.0,64.5192)
        };
        \addplot[color=teal,mark=star,] coordinates {
        (1.0,52.7179)(2.0,58.2179)(3.0,62.1367)(4.0,65.3333)
        };
        \legend{Ours ,SPIn-NeRF ,InFusion ,MVIP-NeRF ,MALD-NeRF}
    \end{axis}
\end{tikzpicture}
    \end{subfigure}
    \vspace{-0.4in}
    \caption{
    Evaluation of 3D consistency of the object removal task on the SPIn-NeRF dataset using TSED.
    }
    \label{fig:tsed-spinnerf}
\end{figure}
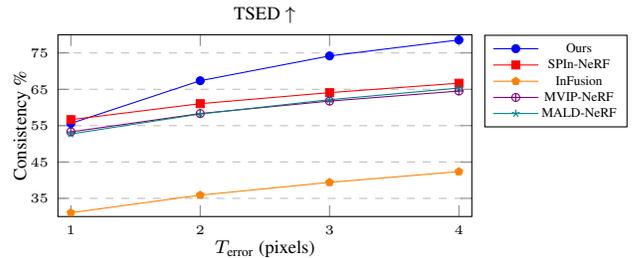

\begin{figure}[t]
    \centering
    \begin{subfigure}{\linewidth}
        \pgfplotsset{width=0.8\linewidth,height=9em,compat=1.18}
\begin{tikzpicture}
    \begin{axis}[
        title style={align=center, font=\scriptsize, yshift=-.5em},
        title={TSED $\uparrow$},
        xlabel={$T_\text{error}$ (pixels)},
        ylabel={Consistency \%},
        xmin=0.9, xmax=4.1,
        ymin=10, ymax=70,
        xtick={1,2,3,4},
        ytick={20,40,60},
        legend pos=outer north east,
        legend style={nodes={scale=0.5, transform shape}},%
        label style={font=\scriptsize},
        tick label style={font=\tiny},
        ymajorgrids=true,
        grid style=dashed,
        xlabel style={yshift=1ex},
        ylabel style={yshift=-1.5ex},
        mark size=1.5pt,x
    ]
        \addplot[color=blue,mark=*,] coordinates {
        (1.0,42.7083)(2.0,52.9167)(3.0,59.3750)(4.0,63.5417)
        };
        \addplot[color=teal,mark=star,] coordinates {
        (1.0,23.1250)(2.0,25.4167)(3.0,27.8472)(4.0,29.4792)
        };
        \addplot[color=violet,mark=oplus,] coordinates {
        (1.0,14.7917)(2.0,18.9583)(3.0,22.1528)(4.0,24.6354)
        };
        \addplot[color=magenta,mark=triangle*,] coordinates {
        (1.0,18.5417)(2.0,22.0833)(3.0,24.8611)(4.0,27.5521)
        };
        \legend{Ours
        ,SPIn-NeRF ,NeRFiller ,NeRFiller w/o depth}
    \end{axis}
\end{tikzpicture}
    \end{subfigure}
    \vspace{-0.4in}
    \caption{
    Evaluation of 3D consistency on the few-view inpainting task.
    }
    \label{fig:tsed-few-view}
\end{figure}
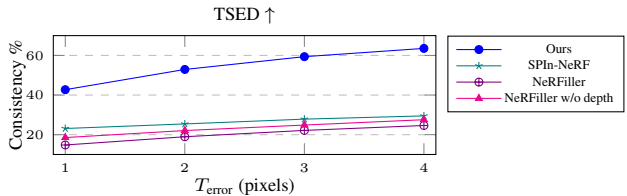

Figures \ref{fig:tsed-spinnerf}, \ref{fig:tsed-nerfiller}, and \ref{fig:tsed-few-view}
provide a comprehensive visualization of TSED results across different tasks and different values of $T_\text{error}$. In \cref{fig:tsed-spinnerf}, we observe that when $T_\text{error} = 1.0 \text{px}$, almost all other methods achieve the same consistency as us, but do not improve (increase) as much as the error threshold increases.
This is primarily because other methods enforce 3D consistency by fusing cross-view information through a 3D radiance field, resulting in blurry output renders.
Since TSED computes SIFT features, naturally fewer such features will be detected from a blurry image, damaging the TSED score. In contrast, our method produces significantly sharper images (see \cref{tab:spinnerf} for details), resulting in more detected features and thus a higher TSED.

In \cref{fig:tsed-nerfiller} (left), when we compare the datasets (i.e., source images, directly from the generative inpainter used in each method), which are used for fitting a NeRF, we observe a significant gap between NeRFiller and ours. 
This is due to our geometry-aware inpainting model,  
which is specifically trained to propagate information across views in a multiview consistent manner.
In contrast, NeRFiller uses a geometry-\textit{un}aware inpainting model, which cannot directly apply the knowledge currently encoded in the 3D scene to inform the inpainting.
Similarly, \cref{fig:tsed-nerfiller} (right) shows that fitting a NeRF on the aforementioned datasets will result in more consistent images overall, but unsurprisingly our consistently inpainted images result in more consistent NeRF renders. Finally, \cref{fig:tsed-few-view} demonstrates the success of our method on inpainting scenes with very few views. We achieve a noticeable improvement over the baselines in terms of TSED consistency, mainly due to the difficulties encountered when fitting NeRFs on very few views, which results in both inconsistency and blurriness.
Please see our webpage for an interactive visualization of SED: \href{https://geomvi.github.io}{https://geomvi.github.io}.

\begin{figure}[t]
    \centering
    \begin{subfigure}{0.34\linewidth}
        \pgfplotsset{width=1.2\linewidth,height=3.2cm,compat=1.18}
\begin{tikzpicture}
    \begin{axis}[
        title style={align=center, font=\scriptsize, yshift=-.5em},
        title={TSED on dataset. $\uparrow$},
        xlabel={$T_\text{error}$ (pixels)},
        ylabel={Consistency \%},
        xmin=0.9, xmax=4.1,
        ymin=5, ymax=80,
        xtick={1,2,3,4},
        ytick={15,35,55,75},
        legend pos=outer north east,
        legend style={nodes={scale=0.5, transform shape}},
        label style={font=\scriptsize},
        tick label style={font=\tiny},
        ymajorgrids=true,
        grid style=dashed,
        xlabel style={yshift=1ex},
        ylabel style={yshift=-1ex},
        mark size=1.5pt,x
    ]
        \addplot[color=blue,mark=*,] coordinates {
        (1.0,56.1709)(2.0,67.8006)(3.0,73.4705)(4.0,77.1361)
        };
        \addplot[color=red,mark=square*,] coordinates {
        (1.0,9.8101)(2.0,16.5348)(3.0,22.4156)(4.0,27.5712)
        };
        \addplot[color=teal,mark=star,] coordinates {
        (1.0,8.5443)(2.0,14.6361)(3.0,20.5169)(4.0,25.7516)
        }; 
        \addplot[color=orange,mark=pentagon*,] coordinates {
        (1.0,8.2278)(2.0,9.8101)(3.0,11.0232)(4.0,12.1440)
        }; 
    \end{axis}
\end{tikzpicture}
    \end{subfigure}
    \begin{subfigure}{0.60\linewidth}
        \pgfplotsset{width=0.72\linewidth,height=3.2cm,compat=1.18}
\begin{tikzpicture}
    \begin{axis}[
        title style={align=center, font=\scriptsize, yshift=-.5em},
        title={TSED on NeRF renders. $\uparrow$},
        xlabel={$T_\text{error}$ (pixels)},
        xmin=0.9, xmax=4.1,
        ymin=85, ymax=100,
        xtick={1,2,3,4},
        ytick={85,90,95,100},
        legend pos=outer north east,
        legend style={nodes={scale=0.5, transform shape}},
        label style={font=\scriptsize},
        tick label style={font=\tiny},
        ymajorgrids=true,
        grid style=dashed,
        xlabel style={yshift=1ex},
        ylabel style={yshift=-1.5ex},
        mark size=1.5pt,x
    ]
        \addplot[color=blue,mark=*,] coordinates {
        (1.0,98.1013)(2.0,98.2595)(3.0,98.4177)(4.0,98.4968)
        };
        \addplot[color=red,mark=square*,] coordinates {
        (1.0,95.8861)(2.0,96.0443)(3.0,96.0970)(4.0,96.1234)
        };
        \addplot[color=teal,mark=star,] coordinates {
        (1.0,93.3544)(2.0,93.5127)(3.0,93.5654)(4.0,93.6709)
        }; 
        \addplot[color=orange,mark=pentagon*,] coordinates {
        (1.0,85.1266)(2.0,86.5506)(3.0,87.5527)(4.0,88.2911)
        }; 
        \legend{Ours ,NeRFiller ,NeRFiller w/o depth, Stable Diffusion (2D)}
    \end{axis}
\end{tikzpicture}
    \end{subfigure}
    \vspace{-0.25in}
    \caption{
    Evaluation of 3D consistency of the scene completion task on the NeRFiller dataset using TSED. 
    The left inset shows the consistency of the ``source images'' (dataset) used to train the NeRF (i.e., the direct outputs of the generative inpainter used in each method). For NeRFiller, we use the dataset from the latest Dataset Update iteration.
    The right inset shows the consistency of the NeRF renders, from a NeRF fit to those source images.
    }
    \label{fig:tsed-nerfiller}
\end{figure}
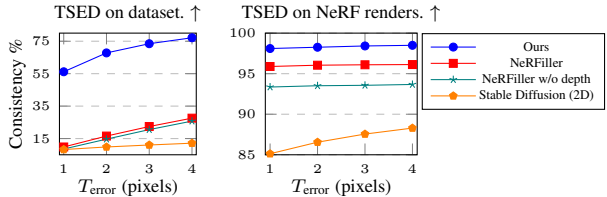

\section{Comprehensive Ablation Studies}
\label{supp:sec:full-ablation}

We extend the ablation studies presented in \S\ref{subsec:ablation} along three key axes: (i) comparison with the naive baseline of independent 2D inpainting, (ii) ablation of inference-time strategies, and (iii) ablating or varying various design choices in our training and fusion strategies.

\subsection{Comparison with Independent 2D inpainting}
\label{supp:subsec:independent-ablation}

\begin{table}[!t]
\centering
\tablesize
\setlength{\tabcolsep}{0.5em}
\begin{tabular}{c|ccc|cc}
Method            & PSNR $\uparrow$               & SSIM $\uparrow$              & LPIPS $\downarrow$           & MUSIQ $\uparrow$             & Corrs $\uparrow$ \\ \hline
Stable Diffusion (2D) \cite{stable.diffusion} & 24.69	          & 0.85	          & 0.10	          & 3.77	            & 1120 \\
ControlNet (2D) \cite{zhang2023controlnet}    & 21.33	          & 0.83	          & 0.14	          & 3.70	            & 1024 \\
BrushNet (2D) \cite{ju2024brushnet}           & 22.84	          & 0.83	          & 0.13	          & 3.77	            & 1081 \\
Ours                                         & \rankonecolor28.59 & \rankonecolor0.89 & \rankonecolor0.05 & \rankonecolor3.80 & \rankonecolor1250    \\
\end{tabular}
\caption{
    Evaluating our method against independent inpainting on scene completion. ``2D'' indicates independent 2D inpainting.
}
\label{tab:independent-ablation}
\end{table}

As observed in prior work \cite{weber2024nerfiller}, independent inpainting fails to produce consistent content across views.
Since \emph{reference-based geometry-awareness} is one of the core components of our approach, we also present a comparison between our geometry-aware inpainting and the naive baseline of geometry-\emph{un}aware inpainting (i.e., independent 2D inpainting), in \cref{tab:independent-ablation}.
We compare our approach to three state-of-the-art diffusion-based 2D inpainting methods: Stable Diffusion \cite{stable.diffusion} (which our model is based on), ControlNet \cite{zhang2023controlnet}, and BrushNet \cite{ju2024brushnet}.
Note that our model, just as for the 2D inpainter baselines, is also a latent diffusion model with a similar architecture, operating on a single image at a time. In other words, we do not use an explicit or implicit 3D radiance field when inpainting the views. The difference lies only in the conditioning signals: our diffusion model is informed by the 3D world and other views through the various cues passed to the generator at inference time.
After inpainting all the views, similar to our method, a NeRF is fit to the inpainted views and the rendered images and videos are assessed to compute the evaluation metrics (refer to \S\ref{subsec:eval-protocol} for details). As demonstrated, our method significantly outperforms all baselines, confirming its superior quality in the context of 3D inpainting.

\subsection{Ablation of Inference-Time Strategies}

\begin{table}[t]
\centering
\tablesize
\setlength{\tabcolsep}{0.8em}
\begin{tabular}{ccc|ccc|cc}
$\Pi$       & $D$        & St. & PSNR $\uparrow$               & SSIM $\uparrow$              & LPIPS $\downarrow$           & MUSIQ $\uparrow$             & Corrs $\uparrow$             \\ \hline
\Checkmark   & \Checkmark   & SR &                27.42 &                0.88 & 0.07                & 3.77                & 1232 \\
\XSolidBrush & \XSolidBrush & AR &                28.32 &                0.88 & \rankonecolor{0.05} & 3.78                & 1235 \\
\XSolidBrush & \Checkmark   & AR & 28.29                &                0.88 & \rankonecolor{0.05} & 3.79                & 1231 \\
\Checkmark   & \XSolidBrush & AR & \ranktwocolor{28.44} & \rankonecolor{0.89} & \rankonecolor{0.05} & \rankonecolor{3.80} & \rankonecolor{1252} \\
\Checkmark   & \Checkmark   & AR & \rankonecolor{28.59} & \rankonecolor{0.89} & \rankonecolor{0.05} & \rankonecolor{3.80} & \ranktwocolor{1250} \\
\end{tabular}%
\caption{Ablation of available inputs and inpainting strategies on the NeRFiller scenes dataset. We denote the camera parameters as $\Pi$, depth maps as $D$, inpainting strategy as St., single-reference inpainting as SR, and autoregressive inpainting as AR. Note that the last row represents our full strategy.}
\label{tab:nerfiller-ablation}
\end{table}

\begin{figure}[t]
    \centering
    \includegraphics[width=\linewidth]{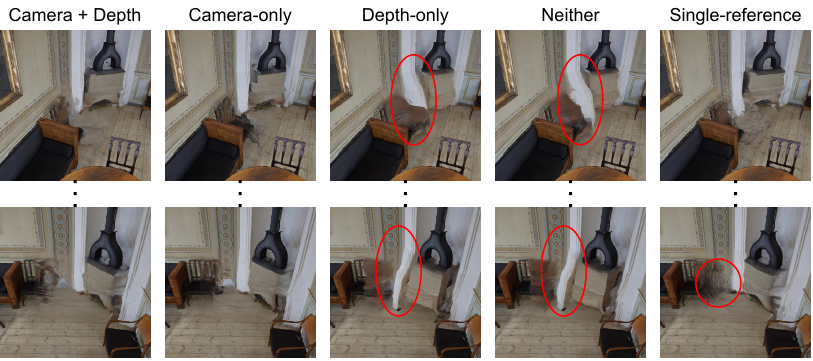}
    \caption{
    A qualitative example from the ablation of inference-time strategies, confirming ground-truth camera parameters and the autoregressive procedure affect the quality of the inpainted scene, whereas ground-truth depth maps have little impact.
    }
    \label{fig:ablation-qualitative}
\end{figure}

\begin{table*}[!th]
\centering
\tablesize
\begin{tabular}{ccccc|ccc|cc}
Training datasets & $\mathcal{G_\mathcal{R}}$ & Reference & Perturbation & Fusion            & PSNR $\uparrow$               & SSIM $\uparrow$              & LPIPS $\downarrow$           & MUSIQ $\uparrow$             & Corrs $\uparrow$ \\ \hline
RealEstate10K + GSO  & \XSolidBrush   & Naive                  & N/A               & Weighted average  &
24.34                & 0.85              & 0.10              & \ranktwocolor3.79              & 1113       \\
RealEstate10K + GSO  & \Checkmark     & Reprojected           & \Checkmark        & Hierarchical      &
27.12	             & 0.87	             & 0.06	             & 3.76  
                     & 1182        \\
COCO + GSO        & \XSolidBrush   & Reprojected           & \Checkmark        & Single confidence &
28.36                & 0.88              & \rankonecolor0.05 & \rankthreecolor3.78 & \ranktwocolor1223    \\
COCO + GSO        & \Checkmark     & Reprojected           & \Checkmark        & Weighted average  &
\rankonecolor29.45   & \rankonecolor0.89 & 0.06              & 3.77              & 1166    \\
COCO + GSO        & \Checkmark     & Reprojected           & \Checkmark        & Closest camera    &
27.36                & 0.88              & 0.06              & \rankthreecolor3.78 & 1204    \\
COCO + GSO        & \Checkmark     & Reprojected           & \XSolidBrush      & Hierarchical      &
\ranktwocolor29.25   & \rankonecolor0.89 & \rankonecolor0.05 & 3.72              & \rankthreecolor1222    \\
COCO + GSO        & \Checkmark     & Reprojected           & \Checkmark        & Hierarchical      &
\rankthreecolor28.59 & \rankonecolor0.89 & \rankonecolor0.05 & \rankonecolor3.80 & \rankonecolor1250    \\
\end{tabular}
\caption{Ablation of various design choices in training and fusion, including conditioning signals, datasets, and other algorithmic components. We denote the presence of geometric cues in the conditioning signals as $\mathcal{G_\mathcal{R}}$. `Closest camera' and `Weighted average' ignore the predicted confidence masks; the former solely conditions on the closest view that has been inpainted, and the latter takes a weighted average of the noise estimates, proportional to the inverse view distance between the reference view, $r$, and the target view, $t$, i.e., $\frac{1}{d((r, t))}$ (\cref{eq:view-distance}). `Single confidence' means that, since back-face and shadow masks are disabled, there is only one confidence signal, derived from the front-face mask. The last row represents our full model, which is superior on nearly all metrics, compared to other variants.
}
\label{tab:model-ablation}
\end{table*}

In \cref{tab:nerfiller-ablation}, 
we ablate various inference-time strategies of our inpainting pipeline on scene completion. 
We observe that, for wide-baseline datasets like NeRFiller, our autoregressive procedure (\S\ref{subsec:autoregressive}) is essential, as a single reference lacks sufficient information for a wide baseline (first row).
We also find that providing \duster with ground-truth depth maps has little effect on performance, highlighting the robustness of our method. However, ground-truth camera parameters have a more significant impact (second to fourth rows). This is mainly because optimizing camera parameters in \duster involves complex global alignment, whereas, when camera parameters are known, optimizing the depth maps becomes a much simpler task.
The qualitative example in \cref{fig:ablation-qualitative} also confirms our findings.

\subsection{Model Design Ablation}

Finally, we ablate or vary several design decisions in our training and fusion strategies, including training datasets, availability of geometric cues ($\mathcal{G_\mathcal{R}}$) in the conditioning signals, whether to align the reference images to the coordinate of the target image by reprojection, and whether to use mesh perturbation. According to \cref{tab:model-ablation}, we find that although mesh perturbation (\S\ref{subsec:inptrain}) does not improve image-based metrics, it has a significant impact on video-based metrics, i.e., a higher image quality and consistency.

Moreover, we find that it is essential to use our hierarchical fusion method (\S\ref{subsubsec:geo-aware-inpainting}), as alternative approaches such as ``weighted average'' and ``closest camera'' lead to lower performance. On the other hand, ``weighted average'' achieves the highest PSNR and SSIM among all settings. 
This is primarily because averaging multiple noise estimates may fuse inconsistent information, producing a blur artifact similar to NeRF renders. Since the inpainted images are already significantly blurred, fitting a NeRF does not introduce additional blurriness. This will result in higher consistency between the inpainted images and their corresponding NeRF renders, leading to higher PSNR and SSIM, though at the cost of lower overall image quality, as reflected in other metrics.

We also observe the importance of conditioning the inpainter on the geometric cues (\S\ref{subsubsec:cond-inp}).
Note that in this case, fusion is performed using a single confidence mask; with back-face and shadow masks disabled, the only confidence signal comes from the front-face mask.

As mentioned in \S\ref{subsec:inptrain}, we use a single-view image dataset instead of a multiview one, to ensure a greater data diversity. To explore the effectiveness of a single-view dataset, we compare our base model with the same model trained on RealEstate10K \cite{realestate10k} instead of COCO \cite{coco.dataset}. 
RealEstate10K, which includes a large set of scenes represented as posed multiview image sets, is commonly used for training large-scale cross-dataset novel view synthesis models (e.g., \cite{wang2021ibrnet,yu2024polyoculus}). 
To obtain multiview depth maps for RealEstate10K, we run \duster on all the videos as a pre-processing stage. \cref{tab:model-ablation} shows that our base model outperforms the one trained on RealEstate10K.

Finally, we evaluate a model naively conditioned on the reference image without any 3D reprojection, instead of our coordinate-aligned conditional inpainting. As reference-based photometric and geometric cues are synthesized for a single-view image dataset like COCO, and no actual reference image exists, we train this model on RealEstate10K.
In other words, for naive conditioning,
we must use a dataset with multiple posed views per scene,
so that one may be used as target and the other as conditioning;
our synthetic single-image reprojections, which have only one frame, therefore cannot be used. 
As presented in \cref{tab:model-ablation} (last two rows), coordinate-aligned conditional inpainting outperforms naive conditioning on RealEstate10K.

\section{Limitations}
\label{supp:sec:limitations}

\begin{figure}[t]
    \centering
    \includegraphics[width=\linewidth]{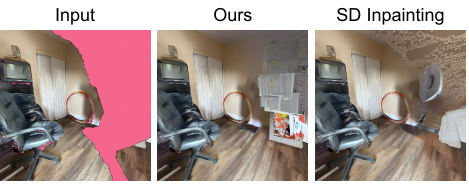}
    \caption{Our inpainting model inherits the limitations of Stable Diffusion for inpainting.}
    \label{fig:sd-limitation}
\end{figure}

While we have demonstrated improved image quality and cross-view consistency over existing baselines, in addition to applicability to the few-view scenario, some shortcomings remain in our approach. Our method relies on two external tools, Stable Diffusion \cite{stable.diffusion} for inpainting and \duster \cite{dust3r} for geometry estimation. Hence, errors caused by these tools (e.g., highly implausible inpaintings or errors in depth estimation) will be propagated throughout the process, resulting in degraded outcomes. In the case of extreme failure in these external tools (e.g. extreme errors in camera and depth estimation), our method is unable to recover.
For instance, consider \cref{fig:sd-limitation}, showing the \emph{initial} view for inpainting the ``office'' scene from the NeRFiller dataset. Clearly, the failure of our method is inherited from the Stable Diffusion inpainter, likely caused by out-of-distribution conditioning (highly irregular inpainting mask in this case). Nevertheless, our method does not catastrophically fail in such cases, even as extreme as this one, maintaining geometry and appearance quality elsewhere in the scene.
We also expect that future improvements to the mentioned tools will also be imparted to our approach.

\end{document}